\documentclass[preprint]{revtex4-2}
    
\usepackage{graphicx}
\usepackage{siunitx}
\usepackage{hyperref}
\usepackage{gensymb}
\usepackage{float}
\usepackage{lineno}
\usepackage{xcolor}

\graphicspath{{./figures_v05/}}

\begin{document}
    \title
    {
    Tracking perovskite crystallization via deep learning-based feature detection on 2D X-ray scattering data
    }
    
    \author{Vladimir Starostin\textsuperscript{a}}
    \author{Valentin Munteanu\textsuperscript{a}}
    \author{Alessandro Greco\textsuperscript{a}}
    \author{Ekaterina Kneschaurek\textsuperscript{a}}
    \author{Alina Pleli\textsuperscript{a}}
    \author{Florian Bertram\textsuperscript{b}}
    \author{Alexander Gerlach\textsuperscript{a}}
    \author{Alexander Hinderhofer\textsuperscript{a}}
    \author{Frank Schreiber\textsuperscript{a}}
    \affiliation{
    [a] Institute of Applied Physics, University of Tübingen, 72076 Tübingen, Germany\\
    [b] Deutsches Elektronen-Synchrotron DESY, Notkestr. 85, 22607 Hamburg, Germany
    }
    
    \begin{abstract}
    
        Understanding the processes of perovskite crystallization is essential for improving the properties of organic solar cells. \textit{In situ} real-time grazing-incidence X-ray diffraction (GIXD) is a key technique for this task, but it produces large amounts of data, frequently exceeding the capabilities of traditional data processing methods. We propose an automated pipeline for the analysis of GIXD images, based on the Faster R-CNN deep learning architecture for object detection, modified to conform to the specifics of the scattering data. The model exhibits high accuracy in detecting diffraction features on noisy patterns with various experimental artifacts. We demonstrate our method on real-time tracking of organic-inorganic perovskite structure crystallization and test it on two applications: 1. the automated phase identification and unit-cell determination of two coexisting phases of Ruddlesden–Popper 2D perovskites, and 2. the fast tracking of MAPbI$_3$ perovskite formation. By design, our approach is equally suitable for other crystalline thin-film materials.
    \end{abstract}
    
    \maketitle

    \renewcommand\thefigure{\arabic{figure}}
    \renewcommand\thetable{\arabic{table}}
    
    \section{Introduction}
    
        Perovskite materials have been heralded as the future of solar cell technology, promising low cost and high efficiency \cite{2009_Miyasaka,2014_perovskite_review}. In the last decade, there has been an enormous interest in enabling the commercial use of perovskite solar cells by optimizing their structural and optoelectronic properties. To this end, a detailed understanding of their crystallization pathways is essential. Grazing-incidence X-ray diffraction (GIXD, which we here consider equivalent to grazing-incidence wide-angle scattering, GIWAXS) is a key technique for this task, enabling real-time, non-destructive probing of perovskite crystal structures during their synthesis \cite{Sinha1988,2017insitu_perovskites,Chen2018,liu2020stabilization,Zhang2021}. Currently, thanks to advances in detector technology, the amount of collected raw data from \textit{in situ} GIXD experiments reaches millions of images per measurement day, far exceeding the capabilities of traditional data processing methods. As a result, only a tiny portion of the collected data is analyzed, so that important processes remain hidden and undiscovered. Therefore, the full potential of the technique is not exploited, and novel, fast technologies for data analysis are required \cite{Wang2018}. 
        
        One of the solutions is to employ machine learning which has already proven advantageous for the analysis of various kinds of scattering data \cite{Greco_2019_JApplCrystallogr,Greco2021,Samarakoon2020,SanchezGonzalez2017,Ziletti2018,Ryan2018,Oviedo2019}. In particular, there is an increasing interest in data-driven approaches for material science, such as automated phase identification and unit-cell determination from one-dimensional X-ray powder diffraction (XRD) measurements \cite{tatlier2011artificial,Lee2020,Ziletti2018,Ryan2018,Oviedo2019}. However, there are no corresponding methods for GIXD analysis yet. The existing machine learning solutions for two-dimensional diffraction data are mainly focused on the preliminary classification of the images \cite{wang2017x,ke2018convolutional,Liu_2019_MRSCommunications} and refining positions of stand-alone peaks for semi-automated data processing \cite{BraggNet2019, braggnn2020}. The more profound tasks, such as phase identification, determination of preferred orientations and unit cell parameters, require careful extraction of all diffraction feature positions. Once the peak positions are established, one can perform a comprehensive analysis of the diffraction patterns autonomously via existing algorithms for peak indexing, matching, unit-cell determination. Therefore, the peak detection procedure is the key bottleneck on the way to automated GIXD analysis. Furthermore, peak detection is also suitable for fast preliminary ``filtering'' of the data during the experiment.
        
        In this work, we implemented a modern two-stage deep learning object detection approach for fast and accurate Debye-Scherrer ring and segment location in GIXD images. As a model, we chose the Faster R-CNN deep learning architecture \cite{fasterrcnn2015,feature_pyramid_networks_2017}, for which we introduced several essential modifications to conform to the specifics of GIXD diffraction data and increase the performance. We trained the model on simulated images via a simulation procedure aimed at reproducing variate scattering backgrounds and experimental artifacts. We note that the translation-invariant Faster R-CNN algorithm detects each peak independently; thus, our approach is material-agnostic by design, unlike in the case of a model trained to extract material-related information from the measured data at a single step.
        
        As a demonstration of the possible applications of our method, we performed an automated analysis on two \textit{in situ} GIXD datasets of organic-inorganic perovskite structures. However, we emphasize that it is, in fact, generally applicable to essentially any crystalline thin-film materials. First, we track the formation of different phases in 2D halide perovskites. Our model detects diffraction reflections, which are then used to identify the perovskite phases and crystal orientations and track the evolution of the lattice parameters over time, which is crucial for the photovoltaic performance. Second, we demonstrate a fast analysis of over a thousand diffraction images from \textit{in situ} measurements of the conversion of methylammonium lead iodide (MAPbI$_3$) from an intermediate phase during spin-coating and annealing. The algorithm detects over 20000 peaks on 1015 images and can pinpoint the start of perovskite formation during the experiment.

    \section{Results}
    
    \subsection{Data-driven model modifications}
    \label{res:data-specifics}
    
\begin{figure}
            \centering
            \includegraphics[width=7.2in]{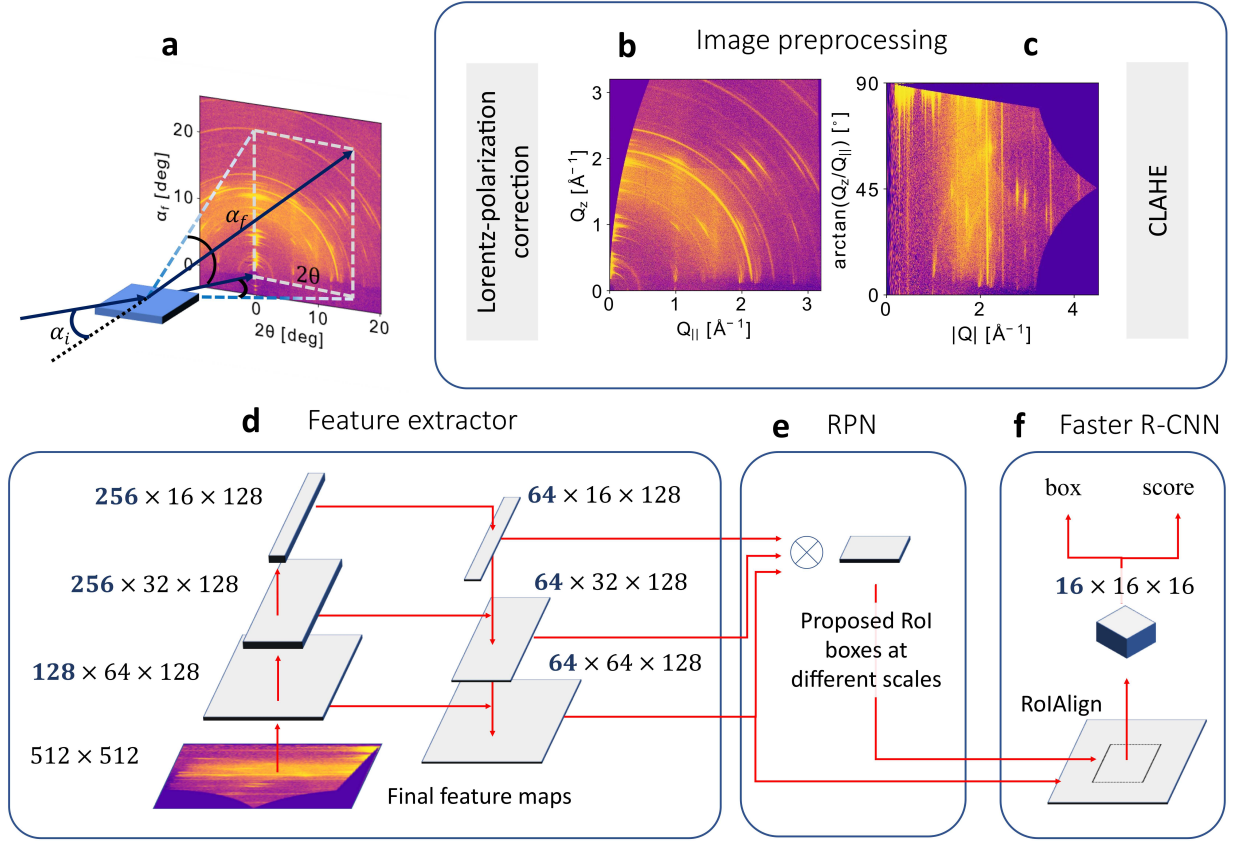}
            \caption{Image preprocessing pipeline and the model architecture. \textbf{(a)} The geometry of the measurements. After Lorentz-polarization correction, measured diffraction patterns are sequentially converted from detector coordinates \textbf{(a)} to reciprocal space \textbf{(b)} and then to polar coordinates \textbf{(c)}. For the detection, the contrast is enhanced by CLAHE (all shown images are already contrast-enhanced for visualization). \textbf{(d)} Feature extractor with asymmetric feature maps; feature shapes correspond to an input image size of $512\times512$ pixels. \textbf{(e)} The Region Proposal Network kernels convolve with feature maps to extract RoI at different scales. \textbf{(f)} At the second detection stage, a RoIAlign layer extracts features at corresponding positions from the largest feature map, from which the box coordinates and the score of confidence are predicted for each box by a fully-connected network.}
            \label{fig:model_scheme}
        \end{figure}

        We implemented a deep learning object detection model based on a two-stage Faster R-CNN framework \cite{fasterrcnn2015,feature_pyramid_networks_2017} to locate diffraction rings and segments in GIXD images. \autoref{fig:model_scheme} illustrates the data pipeline with the model architecture (see Methods \ref{sec:methods:model}). State-of-the-art object detection algorithms are mainly focused on improving performance on objects in RGB photographs and the corresponding benchmark datasets (PASCAL VOC, MS COCO, Open Images, etc.) \cite{fastrcnn2015,fasterrcnn2015,maskrcnn2017,dai2015instanceaware,yolo2016,transformer_detr_2020}. X-ray scattering patterns exhibit particular features that require careful adjustments of the existing general methods. To this end, we introduced an array of modifications to the original architecture to conform to the specifics of the scattering data and accelerate the calculations. In the following, we briefly discuss the key features of the grazing-incidence diffraction images that inform the design of the detection algorithm.

        \paragraph{Polar symmetry.}
        
            X-ray scattering from crystalline grains results in diffraction patterns showing either full rings or ring segments in reciprocal space (depending on their orientation distribution) (see \autoref{fig:model_scheme}b). However, most detection algorithms are designed to detect an object by providing coordinates of a rectangular bounding box around the object. The rectangular shape is most suited for feature extractors that are based on convolution operations with rectangular kernels, while detecting objects with circular shapes would lead to nontrivial complications. To overcome these issues and work with rectangular object shapes, we converted the GIXD images to the polar coordinates $|Q| = ({Q_{z}^2 + Q_{||}^2})^\frac12$ and $\phi = \arctan{(Q_{z}/Q_{||})}$, where $Q$ is the scattering vector and $Q_{||} = ({Q_{x}^2 + Q_{y}^2})^\frac12$. We note that some effects such as refraction may distort the polar symmetry in grazing incidence geometry, but they can be largely neglected here.

        \paragraph{Simple features with complex experimental artifacts.}
        
            In general, Bragg peak profiles can be well approximated by a Voigt function or similar profiles (Gaussian, Pearson VII, pseudo-Voigt, etc. \cite{Rietveld1967,Rietveld1969,David1986}). At the same time, Bragg peaks are typically influenced by various experimental artifacts: incoherent scattering, scattering from amorphous substrates and gases, counting statistics, the direct beam, detector gaps, etc. In this light, we conclude that a deep learning-based solution is required to filter out artifacts and provide stable detection accuracy. However, in contrast to detecting more complex objects in RGB photographs (e.g. humans, animals, cars), a less deep representation might be sufficient for detecting Bragg reflections. Moreover, we can omit a classifier from the detection algorithm, since there is no task to classify peaks by their shape. These simplifications significantly accelerate the model, which is desirable for on-the-fly analysis of measurements with high acquisition rates (up to tens of kHz on modern detectors \cite{EIGER}).

        \paragraph{"Fractal" properties of the detected objects.}
        
            In terms of detection, diffraction rings exhibit properties unusual for objects in RGB photographs. For example, it is possible that a segment of a ring is detected with a higher confidence score than the whole ring. At the same time, two separate but adjacent segments at the same $|Q|$ may be considered a whole ring on a compressed feature map. This behavior causes corresponding detection errors and requires certain adjustments to the training process that are discussed in Methods \ref{sec:methods:model:rpn}, \ref{sec:methods:model:fastrcnn}.
            
        \paragraph{Asymmetric object shapes at different scales.}
            
            Debye-Scherrer rings are typically well-localized in the radial dimension (horizontal in polar coordinates) (see \autoref{fig:model_scheme} (c)). However, their sizes along the angular (vertical) axis may vary substantially depending on the distribution of orientations of the crystalline grains. Modern deep convolutional feature extractors are designed to decrease the spatial resolution in exchange for richer feature maps that may distinguish between different complex shapes. This strategy would be ill-suited for our task as we would prefer to keep a sufficient resolution along the horizontal $|Q|$ axis while compressing the features along the vertical axis. Therefore, we use asymmetric convolutional operations to address the asymmetric shapes of the detected object and to preserve a high resolution along the $|Q|$ axis. We also use several feature maps of different shapes to identify peaks at various scales (see Methods \ref{sec:methods:model:backbone}).
                
        \paragraph{High dynamic range.}
        
            Modern X-ray area detectors can register up to billions of counts per pixel \cite{EIGER} creating exceptionally large dynamic ranges. In practice, such bright spots may correspond to strong reflections from rare large crystalline grains or from the direct beam profile. At the same time, other features may have orders of magnitude less counts, especially at higher $Q$ values. This range in contrast makes it challenging for the neural network to equally detect peaks of different brightness. Therefore, we chose to contrast-enhance the experimental images as a preprocessing step (see Methods \ref{sec:methods:pre-processing}).
        
    \subsection{Performance on the simulated data}
    \label{sec:simulated_results}
        
        \begin{figure}
            \centering
            \includegraphics[width=5in]{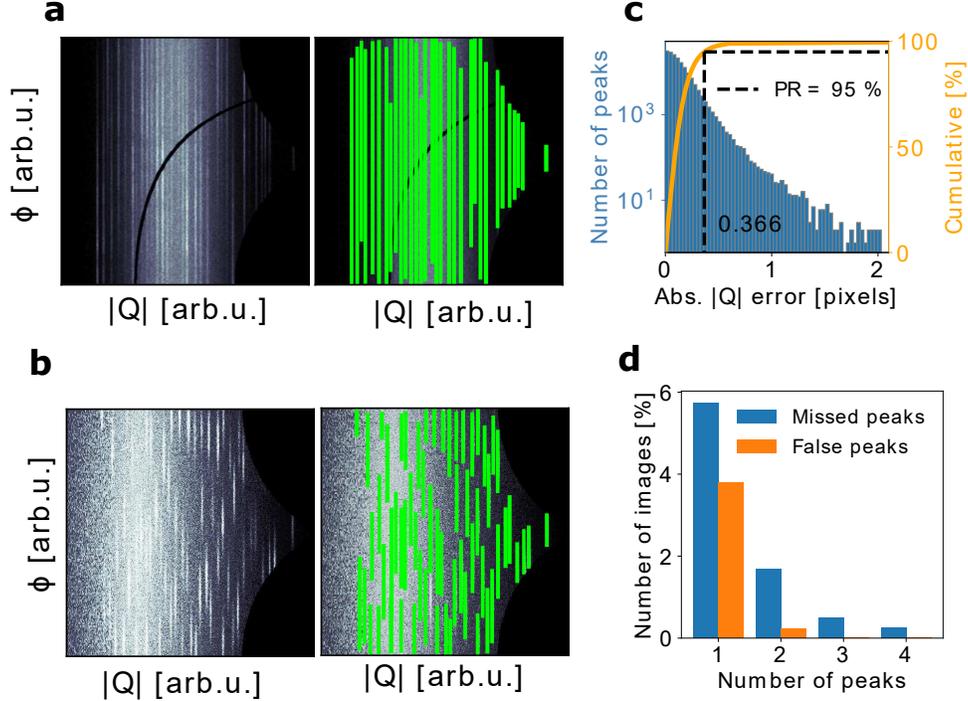}
            \caption{Detection results on the simulated test set of 10000 images. \textbf{(a)} and \textbf{(b)} show two simulated patterns (left) and the detection predictions (right) in the polar coordinates $|Q| = ({Q_z^2 + Q_{||}^2})^{\frac{1}{2}}$ and $\phi = \arctan{(Q_z / Q_{||})}$. \textbf{(c)} The absolute $|Q_\text{detected}|$ error distribution with cumulative curve; the error is below $0.366$ pixels for $95 \%$ of the peaks. \textbf{(d)} Distribution of missed and falsely detected peaks per image.}
            \label{fig:simulation_test_results}
        \end{figure}
    
        The model was trained (see Methods \ref{sec:methods:training}) on a training set of 48000 simulated images which emulate GIXD images in polar space, encompassing a highly diverse configuration of Bragg reflections, scattering backgrounds and experimental artifacts (see Methods \ref{sec:methods:simulations} \textcolor{black}{and Supplementary Figure 1}). The performance of the trained model was first evaluated on a test set of 10000 simulated images. \autoref{fig:simulation_test_results}a-b shows two examples of the simulated patterns. The green lines correspond to the peaks detected by the model. There were $175344$ peaks simulated for 10000 images in total, of which $99.22 \%$ ($173990$) were detected correctly, $0.78 \%$ ($1354$) were missed and $0.25 \%$ ($439$) of the detected peaks were false positives. As a performance metric (\autoref{fig:simulation_test_results}c-d)  we chose to simply use the absolute peak center distance $\Delta|Q| = ||Q_\textrm{detected}| - |Q_\textrm{truth}||$ over the more commonly used metric Intersection over Union (IoU) \cite{fastrcnn2015,fasterrcnn2015,maskrcnn2017,yolo2016}. This was done because we deemed the accuracy of the peak center to be more relevant to this task than the peak overlap. \autoref{fig:simulation_test_results}c shows the distribution of $\Delta|Q|$ for $173990$ peaks along with the cumulative curve; $95 \%$ of the peaks were detected with a $\Delta|Q|$ below $0.366$ pixels. A stochastic nature of the simulation process (see Methods \ref{sec:methods:simulations}) unavoidably creates a small fraction of peaks which are unrecognisable due to noise, background, low intensity, etc.  \autoref{fig:simulation_test_results}d illustrates the distributions of falsely detected and missed peaks per image; only $4 \%$ of images contained at least one falsely detected peak and $8.5 \%$ of images had at least one missed peak. \textcolor{black}{The number of false and missed peaks per image determines the correctness of structure determination. In our case, the vanishingly low numbers of false positive detections per image (0.044 peaks on average) and missed peaks per image (0.14 peaks on average) minimize the corresponding errors in the further analysis of GIXD images.}
        
        \textcolor{black}{We note that our neural network is lightweight (5.9 M parameters) and fast (122 images per second), which is essential for real-time GIXD data analysis. For comparison, the analogous unmodified Faster R-CNN model \cite{feature_pyramid_networks_2017} is substantially larger (41 M parameters) and slower (26 images per second), yet it misses $7 \%$ of the peaks on the test set. This result illustrates that our model is highly optimized for the task and outperforms much larger standard architectures. Furthermore, to justify the choice of the object detection algorithm and the importance of the second detection stage, we provide performance comparisons with two other types of object detection models: the one-stage detector and the segmentation model U-Net \cite{unet_2015}. The results are summarized in Supplementary Figure 2 (see Methods \ref{sec:methods:comparison} for details). Both models exhibit lower detection accuracy than our two-stage detector. Compared to the one-stage detector, the second detection stage is particularly relevant for filtering out false positive predictions (the share of false positives for the one-stage detector is $5.6 \%$). The segmentation model is indeed simpler in terms of implementation; however, in addition to a higher number of false positives ($9.6 \%$), it suffers from other systematic problems: the overlapping peaks cannot be separated, and single peaks can be detected as multiple peaks (see Supplementary Figure 3); the accuracy of peak $|Q|$ determination is limited by pixel size; the process of extracting peak positions from segmentation maps increases the inference time to 30 images per second.}
        
        In the light of these results, it is important to note that, although GIXD data is much harder for humans to interpret than the RGB photographs from typical benchmark object detection datasets, the presented model reached near-optimal performance on the test set. The average precision exceeded $99 \%$, a value which has not yet been reached on datasets with RGB photographs with the current state-of-the-art results being around $60 \%$ \cite{wang2021you}. These results provide evidence that the developed model architecture is well suited for the current task. This success inspired us to apply our strategy to specific scientific problems and automatize the experimental data processing, as explained below.
        
        \begin{figure}[H]
            \centering \includegraphics[width=6.3in]{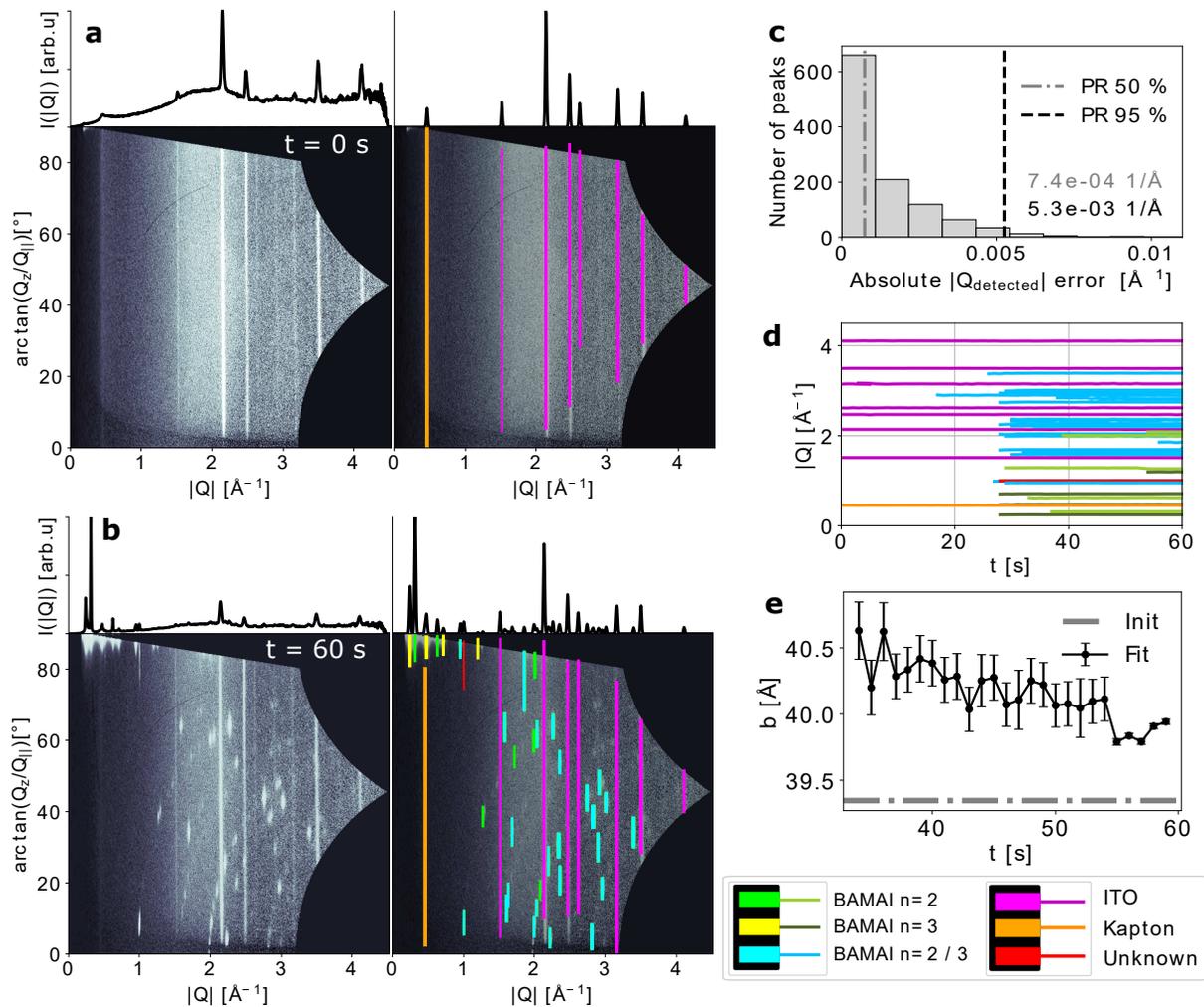}
            \caption{Detection results of \textit{in situ} measurements of 2D halide perovskites formation. Diffraction patterns in \textbf{(a)} and \textbf{(b)} are annotated with Bragg reflections detected by the machine learning algorithm at $t = 0$ s and $t = 60$ s, respectively. Radial profiles above the images show the initial radial profiles calculated from the diffraction patterns (left images) and the profiles calculated based on the detected peaks (right images). \textbf{(c)} The distribution of the absolute error of radial peak positions location by the model with $50 \%$ and $95 \%$ percentiles. \textbf{(d)} The radial positions of the detected peaks over time. \textbf{(e)} The evolution of a lattice parameter $b$ over time for $n=2$ phase of (BA)$_2$(MA)$_{n-1}$Pb$_{n}$I$_{3n+1}$ (denoted as BAMAI on the legend) perovskite calculated via the detected peaks positions.}
            \label{fig:usecase1}
        \end{figure}
        
        \subsection{Use case 1: 2D perovskite lattice parameter         refinement}
        \label{sec:usecase1}

        In the following, we apply the developed model to analyze data from the \textit{in situ} GIXD measurements of the formation of a Ruddlesden–Popper 2D perovskite (butylammonium methylammonium lead iodide (BA)$_2$(MA)$_{n-1}$Pb$_{n}$I$_{3n+1}$  \cite{Stoumpos2013,Stoumpos2016,bamai_n6n7_2018}) during annealing.
        
        The experiment was performed at the synchrotron radiation source PETRA III, at beamline P08 \cite{2011petraIII}. The X-ray beam energy was $E=$ 18\,keV and the angle of incidence was $\alpha_{i}=0.5 \degree $, which is above the critical angle for the substrate $\alpha_c = 0.16 \degree$. The investigated thin film sample was obtained by spin-coating a solution of butylammonium iodide (BAI), methylammonium iodide (MAI) and lead iodide (PbI$_2$) dissolved in a dimethyl formamide and dimethyl sulfoxide mixture (DMF:DMSO = 1:4), on a glass substrate coated with indium tin oxide (ITO) (Supplementary Figure 4). An IR lamp mounted at the top of the spin coating chamber was used to anneal the sample (see Supplementary Figure 5). Sixty diffraction patterns with $1$\,s exposure time each were analyzed, corresponding to 60\,s of annealing.
        
        Figures \ref{fig:usecase1}a-b demonstrate the detection results of the neural network on two diffraction patterns for $t = 0$\,s and $t = 60$\,s. The lines on the right-hand images correspond to the locations of the detected peaks. Our algorithm detected most of the visible reflections - $1156$ diffraction rings and segments from $60$ time frames remained after the filtering stage (see Methods \ref{sec:methods:postprocessing}). The missed reflections typically have low intensities. Some detection inaccuracies in determining peak angular sizes (in particular, lower edges of detected boxes do not cover the whole peaks) may be related to angular profile dissimilarities between the experimental and simulated peaks.
        
        To measure the accuracy of the obtained peak positions, we extracted one-dimensional radial profiles of each peak based on their predicted locations and fitted them with a Gaussian function with linear background via the Levenberg–Marquardt algorithm \cite{lmfit2014}:
        
        $$I(|Q|) = I_0 \exp{(- \frac{(|Q| - Q_\text{fit}) ^ 2}{2 w^ 2})} + B|Q| + C$$
        
        \autoref{fig:usecase1}c shows the distribution of the absolute error of the detected positions $|Q_\text{detected}|$ with respect to $Q_\text{fit}$ obtained by the fit; $95 \%$ of the errors are below $5.3 \times 10^{-3}\si{\angstrom}^{-1}$. The colors of the peaks in \autoref{fig:usecase1}a-b denote the result of the indexing algorithm used for the identification of the Bragg reflections (see the legend on \autoref{fig:usecase1}). We note that our detection algorithm can be coupled with any standard indexing algorithm for diffraction patterns, most (if not all) of which require careful peak position extraction. In this case, it is sufficient to use a simple algorithm for phase identification based on the share of simulated intensities (structure factors) that appear within any of the detected boxes (see Methods \ref{sec:methods:indexing} for more details). We expect some of the possible phases of (BA)$_2$(MA)$_{n-1}$Pb$_{n}$I$_{3n+1}$ to emerge during annealing, where $n \in [1, 7]$ defines the perovskite layer thickness \cite{bamai_n6n7_2018}. Supplementary Figure 7 shows that two phases $n = 2$ and $n = 3$ with $(010)$ orientation are identified by our algorithm. The algorithm also identified 7 peaks from the ITO substrate and one peak from the kapton window. 
        
        Once the phases are identified, each peak is automatically indexed (one peak remains unidentified and likely belongs to a side product of the annealing). \autoref{fig:usecase1}d shows radial peak positions over time, from which we can obtain the moment when both perovskite phases emerge ($t \approx 28$ s). One peak was assigned as belonging to both $n = 2$ and $n = 3$ phases and emerging earlier at $t = 17$ s, however, it is in fact a weak reflection from the ITO substrate at $Q = 2.91\si{\angstrom}^{-1}$.
        
        Supplementary Figure 8 demonstrates a systematic change of the relative radial position of a perovskite peak over time that might indicate changes in the lattice structure. To study these changes, we performed the unit cell refinement procedure (see Methods \ref{sec:methods:unit-cell-refinement}) for both phases for each time frame. \autoref{fig:usecase1}\textcolor{black}{e} demonstrates the behavior of the lattice parameter $b$ of $n=2$ phase that slowly decreases over time from $40.5\si{\angstrom}$, which is about $1\si{\angstrom}$ above the previously reported value \cite{Stoumpos2013} that was used as an initial value for the fitting. The lattice length $b$ of the $n=3$ phase is constant ($b \approx 52.5\si{\angstrom}$) and slightly above the initial value $51.96\si{\angstrom}$. \textcolor{black}{This discrepancy is most likely related to the increased temperature during annealing resulting in asymmetric thermal expansion in the direction corresponding to the spacer molecule.} The other parameters (see Supplementary Figure 9) remain unchanged and are in agreement with the initial values.
        
        Note that while the demonstrated analysis can be easily reviewed and adjusted by an expert at each step if necessary, in general there is no need for a manual input. This way, our detection model allows to automate the essential steps of GIXD data analysis from phase identification and peak indexing to refining the unit cell parameters. Performing the same analysis with conventional tools would require a manual selection and fitting of the peaks for each time frame, which is often so time-consuming that in practice just a small fraction of the obtained reflections is fitted for further analysis. In contrast, our approach potentially allows a more comprehensive analysis already during the experiment.

    \subsection{Use case 2: \textit{in situ} tracking of MAPbI$_3$ perovskite formation}
    \label{sec:usecase2}
        As a second use case for our technique, we analyze 1015 diffraction images from an \textit{in situ} spin-coating and annealing experiment of the formation of 3D MAPbI$_3$ perovskite. The experimental setup was identical to the one described for use case 1. The thin film sample was obtained by spin-coating a solution of MAI and PbI$_2$ dissolved in a mixture of dimethylformamide and dimethyl sulfoxide (DMF:DMSO = 4:1), on a glass substrate coated with ITO.
        
\begin{figure}
            \centering \includegraphics[width=6.3in]{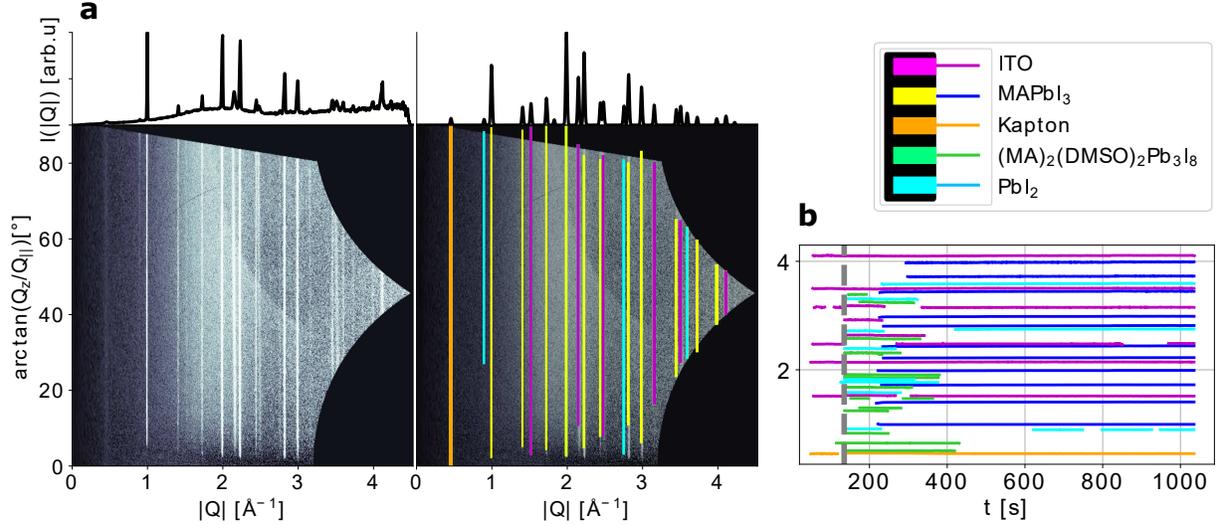}
            \caption{Detection results of \textit{in situ} measurements of MAPbI$_3$ formation during annealing. \textbf{(a)} The diffraction pattern with the detected peaks at $t = 700$ s. \textbf{(b)} The radial positions of the detected peaks over time. The dashed gray line separates the spin-coating and the annealing phases. }
            \label{fig:usecase2}
        \end{figure}

        \autoref{fig:usecase2}a illustrates the diffraction pattern at $t = 700$ s during annealing (left image) and the corresponding detection results (right image). The model detected all the visible peaks in the pattern, which correspond to the reflections from MAPbI$_3$, the ITO substrate, PbI$_2$, the intermediate structure (MA)$_2$(DMSO)$_2$Pb$_3$I$_8$, and the kapton window. The indexing step was performed in the same way as in the previous case \ref{sec:usecase1}. \autoref{fig:usecase2}b shows the $|Q|$ positions of the detected peaks over time. In total, there were 20756 peaks left after the filtering stage (see Methods \ref{sec:methods:postprocessing}). This fast preliminary analysis revealed conversion from the precursor phase (MA)$_2$(DMSO)$_2$Pb$_3$I$_8$ to MAPbI$_3$ perovskite that happens from $t \approx 220$ s to $t \approx 300$ s.

    \section{Discussion}

        We demonstrated an automated deep learning approach for the analysis of GIXD images. We showed how to optimize modern object detection algorithms to address various properties of the experimental data, such as the specific geometry, strong background, asymmetry of the diffraction features, etc. These improvements enabled high performance on diverse experimental data with accurate detection of most of the reflections. The small absolute error of the radial peak position determination and high detection accuracy made it possible to perform an in-depth analysis of \textit{in situ} GIXD data. By processing the extracted feature positions, we identified the formation of two coexisting phases of 2D organic-inorganic perovskite structure and traced the evolution of the lattice parameters in time for these phases. The refinement procedure revealed a slight decrease in the unit cell size for one of the identified phases. We note that such subtle processes may often be overlooked during a manual analysis of the data.
        
        Currently, there is an increasing interest in automated analysis of diffraction data \cite{Huang2021InteractiveVS,Liu_2019_MRSCommunications,Jha2019,Lee2020,Ziletti2018,Ryan2018,Oviedo2019}. An often proposed solution is a single analysis step where the measured raw data is provided to the machine learning algorithm which is supposed to learn which parts of the data carry information, how to perform the desired analysis and obtain material-related results. This approach needs to be tailored as a whole to variations in materials or experimental setups and does not allow to understand or control the process of such an analysis. In contrast, the more flexible analysis approach presented here consists of modular and transparent parts that are simple to control and adjust for a concrete task.
        
        In general, our solution allows to substantially accelerate the analysis process of GIXD images, potentially boosting the speed of scientific discoveries in material science and organic photovoltaics. The other possible applications of the method include the real-time adjustment of the experimental conditions based on the obtained GIXD data and the high-throughput screening of possible perovskite compositions for organic solar cells.

    \section{Methods}
    \label{sec:methods}
    
    \subsection{Image preprocessing}
    \label{sec:methods:pre-processing}
    
        \autoref{fig:model_scheme}a-c demonstrates the individual steps of the preprocessing pipeline applied to all the experimental data used in this work. The measured pattern is first corrected by the Lorentz-polarization factor \cite{Baker2010}, converted to reciprocal space ($Q_{||}, Q_z$) using the knowledge about the experimental geometry and mapped to polar coordinates ($|Q|, \arctan(Q_z / Q_{||})$) with an image resolution $512 \times 1024$ pixels. The image resolution can be extended for larger area detectors without any adjustments to the method. Finally, the images are contrast-enhanced via the contrast-limited adaptive histogram equalization algorithm (CLAHE) \cite{clahe2000}. The contrast-enhanced images are only used for the detection stage, but not for any further analysis.

    \subsection{Model architecture}
    \label{sec:methods:model}
    
        For the detection, we chose a two-stage Faster R-CNN object detection framework \cite{fasterrcnn2015,feature_pyramid_networks_2017,maskrcnn2017} and modified it to meet the specifics of the data discussed above. The general pipeline of the detection process is illustrated by the scheme in \autoref{fig:model_scheme}. First, a convolutional neural network (feature extractor) is applied to an image to convert it into several feature maps with reduced spatial resolution that, in general, carry information about the appearance of different objects on an image. During the first detection stage, a Region Proposal Network (RPN) ``slides'' over the feature maps via the convolution operation. For each pixel on each feature map, it determines whether there is any object at any of the predefined scales (anchors \cite{fasterrcnn2015}) by providing objectness score and the box coordinates relative to the anchors. The resulting regions of interest (RoI) with positive objectness score are individually extracted from the feature maps, reshaped and provided to a fully-connected network \cite{fastrcnn2015} that classifies an object and refines its box coordinates. For more details on the basic implementation we refer to \cite{fasterrcnn2015,feature_pyramid_networks_2017}, and below we focus on the key differences determined by the specifics of the data.
        
        \subsubsection{Feature extractor}
        \label{sec:methods:model:backbone}
            
            We design a feature extractor with asymmetric convolutional layers to preserve resolution along the radial axis $|Q|$. For this purpose, we \textcolor{black}{modify} a small ResNet-18 model pretrained on ImageNet dataset \cite{resnets2016} by using asymmetric stride $= (2, 1)$ for the 3 last residual blocks of the model. \textcolor{black}{The use of the pretrained model allows speeding up the training (see Supplementary Figure 10).} Thus, for an input image of $512\times512$ pixels, the output feature maps from these 3 blocks would have shapes of $64\times128$, $32\times128$ and $16\times128$, respectively. In this way, an image size is mostly reduced in the vertical direction to allow the detection of long narrow rings. At the same time, vertical size reduction from 512 to 16 pixels makes it impossible to resolve segments with small angular (vertical) size. To circumvent this problem, we provide all the 3 feature maps to RPN; each feature map is assigned to extracting objects within a certain size range. As the semantic information is accumulated at the deeper layers of a convolutional network, feature maps from the first blocks provide a shallow representation of an image that in our case might not be sufficient to filter out background and other complex artifacts. To enrich these first feature maps with more complex representation while preserving their resolution, we implement an architecture similar to the Feature Pyramid Network \cite{feature_pyramid_networks_2017}. First, 3 feature maps with sequentially decreasing angular resolution are obtained from the residual blocks as discussed above. After that, smaller and semantically stronger feature maps are summed via upsampling and lateral connections with larger feature maps. To ensure the same number of channels $C = 64$ among the feature maps, the lateral connections are preceded by convolutional layers with $1\times1$ kernels.
            
            It is worth noting that the model can be applied on images of arbitrary size; the shapes discussed above correspond to the simulated images with fixed size $512\times512$ (see Section \ref{sec:methods:simulations}), but for the experimental images the resolution might be increased and it generally depends on the initial resolution of the area detector used in the experiment.
            
        \subsubsection{Region Proposal Network}
        \label{sec:methods:model:rpn}
        
            Our RPN architecture follows the one from \cite{fastrcnn2015} with a reduced number of channels $C = 64$. It is applied to each of the 3 feature maps with the corresponding anchors (see Supplementary Table 1). 
            
           Unlike for usual objects on RGB photographs, there are no well-defined edges of a Gaussian peak; the radial size of a box is arbitrarily defined as a Gaussian RMS width of a peak. However, the surrounding background and overall profile is essential for correct identification of a peak on an image. To address this issue, we extend the target object boxes for RPN by applying ``padding'' $w_{RPN} = 1.1 w_{sim} + 1.5$ pixels, so that the box (therefore, the extracted region on the second detection stage) contains the surrounding background.

        \subsubsection{Second detection stage}
        \label{sec:methods:model:fastrcnn}
        
            To extract a proposed RoI from a feature map, we use RoIAlign layer introduced in \cite{maskrcnn2017} that reshapes a RoI from a feature map into a fixed-shape tensor via bilinear interpolation. Compared to the basic implementation, we increase this resulting shape from $7\times7$ to $16\times16$ to provide a better resolution and, therefore, a better box regression accuracy. At the same time, we decrease the representation size (number of channels) of a RoI from 1024 \cite{fastrcnn2015} to 32.
            
            Unlike usual objects in photographs with complex sharp shapes, several distinct segments at the same $|Q|$ position may be confused with one pronounced segment, especially if a feature map size is drastically reduced along the angular axis. To reduce the probability of this type of detection errors, we only use the first largest feature map for the second detection stage, i.e. RoiAlign layer extracts all the RPN proposals from the largest feature map (see \autoref{fig:model_scheme}). In this way, the use of Feature Pyramid Network is essential to provide maximum information for the second detection stage. 
            
            We reduce the classifier to 2 classes (object / background) and use sigmoid loss function instead of cross-entropy loss just as in RPN classifier. The resulting score of confidence is assigned to refine an objectness score from RPN.

    \subsection{Simulated data for the training}
    \label{sec:methods:simulations}
    
         Training the model on the experimental data would require enormous efforts in annotating large amounts of the measured data manually. Even then there would be a high risk of overfitting on some particular types of data. We find that simulating the data via simple heuristics is sufficient to achieve a very good performance on various types of experimental data. 
         
         When designing the simulation process, we aim at training the model to identify vertical lines with 2D Gaussian profiles on the images superimposed by different experimental artifacts, levels of noise and background signals. An image size of a simulation is $512\times512$ pixels. Most of the simulation stages described below are optional and are invoked with some probability to cover various combinations of the artifacts and features. Moreover, we define probability distributions for most of the simulation parameters so that they differ for every new simulation. The peak positions are stochastic and model-free; the neural network is trained to detect each of the peaks regardless of the position of the other peaks. \textcolor{black}{In this way, the simulations are not limited to a subset of predefined crystal systems; this approach allows to analyze various complex mixed systems with the same neural network for peak detection.}
        
        Each peak profile is modeled by a two-dimensional Gaussian function. To simulate angular profiles of the peaks, which depend on the orientation distribution of the crystalline grains and may exhibit extremely variable patterns, we multiply a simulated map of 2D-Gaussian peaks by Perlin noise \cite{perlin1985}. The simulated background consists of several optional components: linear background, Perlin noise, broad Gaussian profiles. A weighted sum of a background and peak profiles map is further modified by applying Poisson noise to simulate counting statistics, adding detector gaps and geometry-dependent dark areas, etc. \textcolor{black}{The simulation process is organized as a series of sequential image processing steps, and each step adds a certain artifact with a defined probability, so in general each image features multiple artifacts and types of background (see Supplementary Figure 1 and the code for the implementation details).} Finally, simulated images are processed in a similar way as the experimental ones by applying histogram equalization followed by normalization to the range $I \in [0..1]$. \autoref{fig:simulation_test_results} shows several examples of the simulated images.

    \subsection{Training}
    \label{sec:methods:training}
    
        The model was implemented in the Pytorch deep learning framework \cite{Pytorch2019}. We trained both RPN and Fast R-CNN together for 3000 iterations; on each iteration, we simulated a batch of 16 grayscale images with size $512\times512$, in total 48000 images. We used the AdamW optimizer (Adam with decoupled weight decay regularization \cite{loshchilov2019decoupled}) and halved the learning rate every 500th iteration starting from $lr = 0.002$. The loss function contains 4 components:
        
        $$L = \lambda_1 L^{RPN}_{reg} + \lambda_2 L^{RPN}_{score} + \lambda_3 L^{ROI}_{reg} + \lambda_4 L^{ROI}_{score},$$
        where \textit{reg} terms are regression losses for box coordinates calculated as smooth $L1$ loss \cite{fastrcnn2015}, \textit{score} terms are objectness losses calculated as a sigmoid function, $RPN$ and $ROI$ denote the first and the second detection stages, respectively, and $\lambda_i$ are the weights used to balance these terms; we use $\lambda_1  = \lambda_3 = 10$ and $\lambda_2 = \lambda_4 = 1$.
        
        The training data is simulated on the fly during the training. In this way, we did not train a model on a same image twice to eliminate possible overfitting, increase variability of the data, and omit a validation set. The whole training takes less than 25 minutes on a single NVIDIA 2080Ti graphics card. It is worth noting that, since our model is much lighter (i.e. contains much less parameters) than the original Faster R-CNN implementation, it allows to increase batch size from 2 to 16 images per batch and, therefore, unfreeze and train Batch Normalization layers \cite{fastrcnn2015,batch_normalization_2015}. 
    
    \subsection{Postprocessing}
    \label{sec:methods:postprocessing}
        
       One disadvantage of the Faster R-CNN architecture is that a single object can be detected several times with overlapping boxes. Thus, after obtaining the box coordinates and scores of confidence, some of the predictions have to be filtered in order to eliminate possible duplicates. In this work, we use a standard non-maximum suppression operation \cite{nms2006} which removes all overlapping boxes except the one with the highest score of confidence if the degree of overlap exceeds an intersection over union metric \cite{fastrcnn2015} $IoU = 0.1$. The value of $IoU$ is set quite low to ensure minimal appearance of duplicates, but it can be adjusted if many overlapping peaks are expected on a diffraction pattern.
       
       The other necessary filtering stage that was implemented is the removal of predictions with low confidence score. In this work, we use the threshold value $0.8$, though most of the predicted peaks exhibit a confidence score above $0.95$. When detecting the peaks in \textit{in situ} data, it is desirable to identify and ``connect'' Bragg reflections that appear in sequential time frames. This is achieved by calculating $IoU$ between all the detected peaks from adjacent time frames and connecting those pairs of peaks with largest $IoU$ if it is sufficient (we use an arbitrarily chosen value $IoU = 0.3$; the results of the algorithm are not sensitive to this parameter). As a result, we obtain a set of reflections with detected positions over time and total duration of the presence of each peak. Based on this duration, we are able to perform an additional filtering procedure by removing those peaks with small duration as they are likely to be related to experimental artifacts or noise.
    \subsection{Comparison to other object detection models}
    \label{sec:methods:comparison}
    
          \textcolor{black}{
          In addition to our two-stage detector, we trained the other three neural networks and evaluated their performance on the simulated data (see Supplementary Figure 2).}
          
          \textcolor{black}{
          Faster R-CNN model with ResNet-50 and Feature Pyramid Network \cite{feature_pyramid_networks_2017} with 2 object classes (peaks and background) is trained with 4 images per batch due to memory limitations. Therefore, the training is extended to 12000 iterations. 
          }
          
          \textcolor{black}{
          U-Net \cite{unet_2015} is modified by adding Batch Normalization layers and trained to solve a segmentation task with a cross-entropy loss; the separate peak positions are obtained from the segmentation map via the standard algorithm \cite{label_segmentation_2005,scikit_image_2014}. The confidence score of each peak prediction is estimated by the mean value of the segmentation map within the predicted box. Similar to Faster R-CNN, we use 4 images per batch due to memory limitations and 12000 training iterations in total.
          }
          
          \textcolor{black}{
          Our Region Proposal Network (a one-stage detector) is trained separately; the training process is equivalent to our two-stage detector. 
        }
        
    \subsection{Phase identification and indexing}
    \label{sec:methods:indexing}
        
        For a given crystal structure and orientation, we calculate reflection positions and structure factors for all the combinations of miller indices in a wide range $[-20, 20]$, filter them based on $Q_z, Q_{||}$ ranges measured in the experiment, and determine which reflections appear close enough to the detected peaks based on the following conditions: $||Q_\text{detected}| - |Q_\text{sim}|| / w_\text{detected} \leq 1$ and $|\phi_\text{detected} - \phi_\text{sim}| / a_\text{detected} \leq 1$ where $w_\text{detected},  a_\text{detected}$ are radial and angular sizes of the detected peak, respectively, and $\phi = \arctan(Q_z / Q_{||})$. The fraction of the corresponding structure factors gives a metric which is sufficient for finding the best match among a set of phases and orientations (see Supplementary Fig. 4). Since some of the reflections appear at $Q_{||} = 0$ ($\phi = 90^{\circ}$) and their centers are hidden behind a ``missing wedge'', we automatically prolong those peaks that reach the edge of a ``missing wedge'' to $\phi = 90^{\circ}$ to be matched correctly.
        
        \textcolor{black}{The minimal number of detected peaks required to identify the structure correctly depends on the choice of an indexing algorithm. In general, the employed matching algorithm might require substantially more detected peaks to robustly identify the correct structure compared to standard indexing algorithms that may only require three peaks per structure. Nevertheless, the number of peaks detected by the model on the experimental datasets is sufficient to determine every presented structure with the matching algorithm correctly.}
        
        When there are both reflections from crystalline grains with preferred orientations (``short'' segments in angular dimension) and ``long'' rings from random orientations, it is beneficial to separate the predictions to these two groups and index separately. We do so for the analysis of 2D perovskite data by clustering predicted peaks into 2 groups based on the relative angular size of peaks $a_\text{detected} / H(Q_\text{detected})$, where $H(Q_\text{detected})$ is the maximum possible angular size at a given $|Q|$ position. In this case, the reflections from ITO and kapton are separated from the reflections from 2D perovskites and they are indexed separately. 
        
        The indexing stage is rather trivial and it is performed by assigning the closest simulated reflection to each detected peak given the reflection is close enough (see the conditions above) for each of the phases. Thus, one peak may correspond to several overlapping reflections from different phases (see \autoref{fig:usecase1}).

    \subsection{Unit cell refinement}
    \label{sec:methods:unit-cell-refinement}
    
        Lattice parameters of the identified $n = 2$ and $n = 3$ phases of the (BA)$_2$(MA)$_{n-1}$Pb$_{n}$I$_{3n+1}$ 2D perovskite are refined based on the positions of the detected and indexed Bragg reflections. \textcolor{black}{We achieve this by minimizing the mean squared distance between the detected and simulated peak positions with respect to the lattice parameters.} Some of the reflections from $n = 2$ and $n = 3$ phases overlap and are detected as single peaks. Indeed, the corresponding detected peak positions are, in general, biased. To achieve a better accuracy of the lattice parameters determination, we only use those $N$ peaks that do not overlap with peaks from another phase. The refined lattice parameters are obtained by minimizing 
        $$\chi_{N-3}^2(a,b,c) = \sum^{N}_{i=1}\left(\frac{Q^i_\text{detected} - Q^i_\text{sim}(a,b,c, h_i, k_i, l_i)}{\sigma^i_\text{detected}}\right)^2$$
        with respect to the lattice parameters $a, b, c$ for an orthorhombic unit cell (angles $\alpha=\beta=\gamma=\pi / 2$) for each time frame via L-BFGS-B algorithm \cite{L_BFGS_B_1995}; the Miller indices $\{h_i, k_i, l_i\}$ of an $i^\text{th}$ peak are known from the indexing results. The initial values $a = 8.947\si{\angstrom}, b = 39.347\si{\angstrom}, c = 8.8589\si{\angstrom}$ for $n = 2$ and $a = 8.928\si{\angstrom}, b = 51.959\si{\angstrom}, c = 8.878 \si{\angstrom}$ for $n = 3$ are taken from the previously reported structures \cite{Stoumpos2016}. The standard errors are calculated from the inverse hessian matrix; detection errors are majorized by $\sigma^i_\text{detected} = 0.01\si{\angstrom}^{-1}$.
        
    \section{Code availability}
        \textcolor{black}{The code for object detection model and the simulations is published on \href{https://github.com/schreiber-lab/mlgixd}{https://github.com/schreiber-lab/mlgixd}.}

    \section{Data availability}
        The GIXD datasets are available from the corresponding author upon reasonable request.

    \section{Competing interests}
        The authors declare no competing interests.
        
    \section{Author contributions}
        V.S. and A.H. conceived the concept for the work. V.S. wrote the code, designed and trained the neural network. V.M. carried out the conventional analysis of GIXD data and took part in designing the simulations. F.B., E.K., V.M., and A.H carried out the measurements. V.S., V.M., A.Gr., A.P., A.G., and F.S. wrote the paper and contributed to the analysis and interpretation of the results. F.S. supervised the research process. All authors discussed the results and commented on the manuscript.

    \section{Acknowledgments}
        This research is part of a project funded by the German Federal Ministry for Science and Education (BMBF).
        
        We acknowledge DESY (Hamburg, Germany), a member of the Helmholtz Association HGF, for the provision of experimental facilities. Parts of this research were carried out at PETRA III and we would like to thank Chen Shen and Rene Kirchhof for assistance in using photon beamline P08. Beamtime was allocated for proposal II-20190761. This research was also supported in part through the Maxwell computational resources operated at DESY with the assistance of André Rothkirch and Frank Schlünzen.
        
        We acknowledge Laboratory of Photonics and Interfaces at École Polytechnique Fédérale de Lausanne and the Adolphe Merkle Institute of the University of Fribourg and gratefully thank Jovana Milić, Anwar Alanazi and Algirdas Ducinskas for providing the samples.
        
        We thank the Deutsche Forschungsgemeinschaft (DFG) for financial support.
        
        Supported by the German Research Foundation through the Cluster of Excellence “Machine Learning – New Perspectives for Science”.
        
    \bibliography{references}

\begin{thebibliography}{53}%
\makeatletter
\providecommand \@ifxundefined [1]{%
 \@ifx{#1\undefined}
}%
\providecommand \@ifnum [1]{%
 \ifnum #1\expandafter \@firstoftwo
 \else \expandafter \@secondoftwo
 \fi
}%
\providecommand \@ifx [1]{%
 \ifx #1\expandafter \@firstoftwo
 \else \expandafter \@secondoftwo
 \fi
}%
\providecommand \natexlab [1]{#1}%
\providecommand \enquote  [1]{``#1''}%
\providecommand \bibnamefont  [1]{#1}%
\providecommand \bibfnamefont [1]{#1}%
\providecommand \citenamefont [1]{#1}%
\providecommand \href@noop [0]{\@secondoftwo}%
\providecommand \href [0]{\begingroup \@sanitize@url \@href}%
\providecommand \@href[1]{\@@startlink{#1}\@@href}%
\providecommand \@@href[1]{\endgroup#1\@@endlink}%
\providecommand \@sanitize@url [0]{\catcode `\\12\catcode `\$12\catcode
  `\&12\catcode `\#12\catcode `\^12\catcode `\_12\catcode `\%12\relax}%
\providecommand \@@startlink[1]{}%
\providecommand \@@endlink[0]{}%
\providecommand \url  [0]{\begingroup\@sanitize@url \@url }%
\providecommand \@url [1]{\endgroup\@href {#1}{\urlprefix }}%
\providecommand \urlprefix  [0]{URL }%
\providecommand \Eprint [0]{\href }%
\providecommand \doibase [0]{https://doi.org/}%
\providecommand \selectlanguage [0]{\@gobble}%
\providecommand \bibinfo  [0]{\@secondoftwo}%
\providecommand \bibfield  [0]{\@secondoftwo}%
\providecommand \translation [1]{[#1]}%
\providecommand \BibitemOpen [0]{}%
\providecommand \bibitemStop [0]{}%
\providecommand \bibitemNoStop [0]{.\EOS\space}%
\providecommand \EOS [0]{\spacefactor3000\relax}%
\providecommand \BibitemShut  [1]{\csname bibitem#1\endcsname}%
\let\auto@bib@innerbib\@empty
\bibitem [{\citenamefont {Kojima}\ \emph {et~al.}(2009)\citenamefont {Kojima},
  \citenamefont {Teshima}, \citenamefont {Shirai},\ and\ \citenamefont
  {Miyasaka}}]{2009_Miyasaka}%
  \BibitemOpen
  \bibfield  {author} {\bibinfo {author} {\bibfnamefont {A.}~\bibnamefont
  {Kojima}}, \bibinfo {author} {\bibfnamefont {K.}~\bibnamefont {Teshima}},
  \bibinfo {author} {\bibfnamefont {Y.}~\bibnamefont {Shirai}},\ and\ \bibinfo
  {author} {\bibfnamefont {T.}~\bibnamefont {Miyasaka}},\ }\bibfield  {title}
  {\bibinfo {title} {Organometal halide perovskites as visible-light
  sensitizers for photovoltaic cells},\ }\href
  {https://doi.org/10.1021/ja809598r} {\bibfield  {journal} {\bibinfo
  {journal} {J. Am. Chem. Soc.}\ }\textbf {\bibinfo {volume} {131}},\ \bibinfo
  {pages} {6050} (\bibinfo {year} {2009})}\BibitemShut {NoStop}%
\bibitem [{\citenamefont {Green}\ \emph {et~al.}(2014)\citenamefont {Green},
  \citenamefont {Ho-Baillie},\ and\ \citenamefont
  {Snaith}}]{2014_perovskite_review}%
  \BibitemOpen
  \bibfield  {author} {\bibinfo {author} {\bibfnamefont {M.~A.}\ \bibnamefont
  {Green}}, \bibinfo {author} {\bibfnamefont {A.}~\bibnamefont {Ho-Baillie}},\
  and\ \bibinfo {author} {\bibfnamefont {H.~J.}\ \bibnamefont {Snaith}},\
  }\bibfield  {title} {\bibinfo {title} {The emergence of perovskite solar
  cells},\ }\href {https://doi.org/10.1038/nphoton.2014.134} {\bibfield
  {journal} {\bibinfo  {journal} {Nature Photon}\ }\textbf {\bibinfo {volume}
  {8}},\ \bibinfo {pages} {506} (\bibinfo {year} {2014})}\BibitemShut {NoStop}%
\bibitem [{\citenamefont {Sinha}\ \emph {et~al.}(1988)\citenamefont {Sinha},
  \citenamefont {Sirota}, \citenamefont {Garoff},\ and\ \citenamefont
  {Stanley}}]{Sinha1988}%
  \BibitemOpen
  \bibfield  {author} {\bibinfo {author} {\bibfnamefont {S.~K.}\ \bibnamefont
  {Sinha}}, \bibinfo {author} {\bibfnamefont {E.~B.}\ \bibnamefont {Sirota}},
  \bibinfo {author} {\bibfnamefont {S.}~\bibnamefont {Garoff}},\ and\ \bibinfo
  {author} {\bibfnamefont {H.~B.}\ \bibnamefont {Stanley}},\ }\bibfield
  {title} {\bibinfo {title} {X-ray and neutron scattering from rough
  surfaces},\ }\href {https://doi.org/10.1103/PhysRevB.38.2297} {\bibfield
  {journal} {\bibinfo  {journal} {Phys. Rev. B}\ }\textbf {\bibinfo {volume}
  {38}},\ \bibinfo {pages} {2297} (\bibinfo {year} {1988})}\BibitemShut
  {NoStop}%
\bibitem [{\citenamefont {Hu}\ \emph {et~al.}(2017)\citenamefont {Hu},
  \citenamefont {Zhao}, \citenamefont {Wu}, \citenamefont {Gao}, \citenamefont
  {Luo}, \citenamefont {Jiang}, \citenamefont {Zhang}, \citenamefont {Zhu},
  \citenamefont {Schaible}, \citenamefont {Hexemer}, \citenamefont {Wang},
  \citenamefont {Liu}, \citenamefont {Zhang}, \citenamefont {Gr\"{a}tzel},
  \citenamefont {Liu}, \citenamefont {Russell}, \citenamefont {Zhu},\ and\
  \citenamefont {Gong}}]{2017insitu_perovskites}%
  \BibitemOpen
  \bibfield  {author} {\bibinfo {author} {\bibfnamefont {Q.}~\bibnamefont
  {Hu}}, \bibinfo {author} {\bibfnamefont {L.}~\bibnamefont {Zhao}}, \bibinfo
  {author} {\bibfnamefont {J.}~\bibnamefont {Wu}}, \bibinfo {author}
  {\bibfnamefont {K.}~\bibnamefont {Gao}}, \bibinfo {author} {\bibfnamefont
  {D.}~\bibnamefont {Luo}}, \bibinfo {author} {\bibfnamefont {Y.}~\bibnamefont
  {Jiang}}, \bibinfo {author} {\bibfnamefont {Z.}~\bibnamefont {Zhang}},
  \bibinfo {author} {\bibfnamefont {C.}~\bibnamefont {Zhu}}, \bibinfo {author}
  {\bibfnamefont {E.}~\bibnamefont {Schaible}}, \bibinfo {author}
  {\bibfnamefont {A.}~\bibnamefont {Hexemer}}, \bibinfo {author} {\bibfnamefont
  {C.}~\bibnamefont {Wang}}, \bibinfo {author} {\bibfnamefont {Y.}~\bibnamefont
  {Liu}}, \bibinfo {author} {\bibfnamefont {W.}~\bibnamefont {Zhang}}, \bibinfo
  {author} {\bibfnamefont {M.}~\bibnamefont {Gr\"{a}tzel}}, \bibinfo {author}
  {\bibfnamefont {F.}~\bibnamefont {Liu}}, \bibinfo {author} {\bibfnamefont
  {T.~P.}\ \bibnamefont {Russell}}, \bibinfo {author} {\bibfnamefont
  {R.}~\bibnamefont {Zhu}},\ and\ \bibinfo {author} {\bibfnamefont
  {Q.}~\bibnamefont {Gong}},\ }\bibfield  {title} {\bibinfo {title} {In situ
  dynamic observations of perovskite crystallisation and microstructure
  evolution intermediated from [{PbI}6]4- cage nanoparticles},\ }\bibfield
  {journal} {\bibinfo  {journal} {Nature Communications}\ }\textbf {\bibinfo
  {volume} {8}},\ \href {https://doi.org/10.1038/ncomms15688}
  {10.1038/ncomms15688} (\bibinfo {year} {2017})\BibitemShut {NoStop}%
\bibitem [{\citenamefont {Chen}\ \emph {et~al.}(2018)\citenamefont {Chen},
  \citenamefont {Shiu}, \citenamefont {Ma}, \citenamefont {Alpert},
  \citenamefont {Zhang}, \citenamefont {Foley}, \citenamefont {Smilgies},
  \citenamefont {Lee},\ and\ \citenamefont {Choi}}]{Chen2018}%
  \BibitemOpen
  \bibfield  {author} {\bibinfo {author} {\bibfnamefont {A.~Z.}\ \bibnamefont
  {Chen}}, \bibinfo {author} {\bibfnamefont {M.}~\bibnamefont {Shiu}}, \bibinfo
  {author} {\bibfnamefont {J.~H.}\ \bibnamefont {Ma}}, \bibinfo {author}
  {\bibfnamefont {M.~R.}\ \bibnamefont {Alpert}}, \bibinfo {author}
  {\bibfnamefont {D.}~\bibnamefont {Zhang}}, \bibinfo {author} {\bibfnamefont
  {B.~J.}\ \bibnamefont {Foley}}, \bibinfo {author} {\bibfnamefont {D.-M.}\
  \bibnamefont {Smilgies}}, \bibinfo {author} {\bibfnamefont {S.-H.}\
  \bibnamefont {Lee}},\ and\ \bibinfo {author} {\bibfnamefont {J.~J.}\
  \bibnamefont {Choi}},\ }\bibfield  {title} {\bibinfo {title} {Origin of
  vertical orientation in two-dimensional metal halide perovskites and its
  effect on photovoltaic performance},\ }\bibfield  {journal} {\bibinfo
  {journal} {Nature Communications}\ }\textbf {\bibinfo {volume} {9}},\ \href
  {https://doi.org/10.1038/s41467-018-03757-0} {10.1038/s41467-018-03757-0}
  (\bibinfo {year} {2018})\BibitemShut {NoStop}%
\bibitem [{\citenamefont {Liu}\ \emph {et~al.}(2020{\natexlab{a}})\citenamefont
  {Liu}, \citenamefont {Akin}, \citenamefont {Hinderhofer}, \citenamefont
  {Eickemeyer}, \citenamefont {Zhu}, \citenamefont {Seo}, \citenamefont
  {Zhang}, \citenamefont {Schreiber}, \citenamefont {Zhang}, \citenamefont
  {Zakeeruddin} \emph {et~al.}}]{liu2020stabilization}%
  \BibitemOpen
  \bibfield  {author} {\bibinfo {author} {\bibfnamefont {Y.}~\bibnamefont
  {Liu}}, \bibinfo {author} {\bibfnamefont {S.}~\bibnamefont {Akin}}, \bibinfo
  {author} {\bibfnamefont {A.}~\bibnamefont {Hinderhofer}}, \bibinfo {author}
  {\bibfnamefont {F.~T.}\ \bibnamefont {Eickemeyer}}, \bibinfo {author}
  {\bibfnamefont {H.}~\bibnamefont {Zhu}}, \bibinfo {author} {\bibfnamefont
  {J.-Y.}\ \bibnamefont {Seo}}, \bibinfo {author} {\bibfnamefont
  {J.}~\bibnamefont {Zhang}}, \bibinfo {author} {\bibfnamefont
  {F.}~\bibnamefont {Schreiber}}, \bibinfo {author} {\bibfnamefont
  {H.}~\bibnamefont {Zhang}}, \bibinfo {author} {\bibfnamefont {S.~M.}\
  \bibnamefont {Zakeeruddin}}, \emph {et~al.},\ }\bibfield  {title} {\bibinfo
  {title} {Stabilization of highly efficient and stable phase-pure {FAPbI3}
  perovskite solar cells by molecularly tailored 2d-overlayers},\ }\href@noop
  {} {\bibfield  {journal} {\bibinfo  {journal} {Angewandte Chemie
  International Edition}\ }\textbf {\bibinfo {volume} {59}},\ \bibinfo {pages}
  {15688} (\bibinfo {year} {2020}{\natexlab{a}})}\BibitemShut {NoStop}%
\bibitem [{\citenamefont {Zhang}\ \emph {et~al.}(2021)\citenamefont {Zhang},
  \citenamefont {Eickemeyer}, \citenamefont {Zhou}, \citenamefont
  {Mladenovi{\'{c}}}, \citenamefont {Jahanbakhshi}, \citenamefont {Merten},
  \citenamefont {Hinderhofer}, \citenamefont {Hope}, \citenamefont {Ouellette},
  \citenamefont {Mishra}, \citenamefont {Ahlawat}, \citenamefont {Ren},
  \citenamefont {Su}, \citenamefont {Krishna}, \citenamefont {Wang},
  \citenamefont {Dong}, \citenamefont {Guo}, \citenamefont {Zakeeruddin},
  \citenamefont {Schreiber}, \citenamefont {Hagfeldt}, \citenamefont {Emsley},
  \citenamefont {Rothlisberger}, \citenamefont {Mili{\'{c}}},\ and\
  \citenamefont {Gr\"{a}tzel}}]{Zhang2021}%
  \BibitemOpen
  \bibfield  {author} {\bibinfo {author} {\bibfnamefont {H.}~\bibnamefont
  {Zhang}}, \bibinfo {author} {\bibfnamefont {F.~T.}\ \bibnamefont
  {Eickemeyer}}, \bibinfo {author} {\bibfnamefont {Z.}~\bibnamefont {Zhou}},
  \bibinfo {author} {\bibfnamefont {M.}~\bibnamefont {Mladenovi{\'{c}}}},
  \bibinfo {author} {\bibfnamefont {F.}~\bibnamefont {Jahanbakhshi}}, \bibinfo
  {author} {\bibfnamefont {L.}~\bibnamefont {Merten}}, \bibinfo {author}
  {\bibfnamefont {A.}~\bibnamefont {Hinderhofer}}, \bibinfo {author}
  {\bibfnamefont {M.~A.}\ \bibnamefont {Hope}}, \bibinfo {author}
  {\bibfnamefont {O.}~\bibnamefont {Ouellette}}, \bibinfo {author}
  {\bibfnamefont {A.}~\bibnamefont {Mishra}}, \bibinfo {author} {\bibfnamefont
  {P.}~\bibnamefont {Ahlawat}}, \bibinfo {author} {\bibfnamefont
  {D.}~\bibnamefont {Ren}}, \bibinfo {author} {\bibfnamefont {T.-S.}\
  \bibnamefont {Su}}, \bibinfo {author} {\bibfnamefont {A.}~\bibnamefont
  {Krishna}}, \bibinfo {author} {\bibfnamefont {Z.}~\bibnamefont {Wang}},
  \bibinfo {author} {\bibfnamefont {Z.}~\bibnamefont {Dong}}, \bibinfo {author}
  {\bibfnamefont {J.}~\bibnamefont {Guo}}, \bibinfo {author} {\bibfnamefont
  {S.~M.}\ \bibnamefont {Zakeeruddin}}, \bibinfo {author} {\bibfnamefont
  {F.}~\bibnamefont {Schreiber}}, \bibinfo {author} {\bibfnamefont
  {A.}~\bibnamefont {Hagfeldt}}, \bibinfo {author} {\bibfnamefont
  {L.}~\bibnamefont {Emsley}}, \bibinfo {author} {\bibfnamefont
  {U.}~\bibnamefont {Rothlisberger}}, \bibinfo {author} {\bibfnamefont {J.~V.}\
  \bibnamefont {Mili{\'{c}}}},\ and\ \bibinfo {author} {\bibfnamefont
  {M.}~\bibnamefont {Gr\"{a}tzel}},\ }\bibfield  {title} {\bibinfo {title}
  {Multimodal host{\textendash}guest complexation for efficient and stable
  perovskite photovoltaics},\ }\bibfield  {journal} {\bibinfo  {journal}
  {Nature Communications}\ }\textbf {\bibinfo {volume} {12}},\ \href
  {https://doi.org/10.1038/s41467-021-23566-2} {10.1038/s41467-021-23566-2}
  (\bibinfo {year} {2021})\BibitemShut {NoStop}%
\bibitem [{\citenamefont {Wang}\ \emph {et~al.}(2018)\citenamefont {Wang},
  \citenamefont {Steiner},\ and\ \citenamefont {Sepe}}]{Wang2018}%
  \BibitemOpen
  \bibfield  {author} {\bibinfo {author} {\bibfnamefont {C.}~\bibnamefont
  {Wang}}, \bibinfo {author} {\bibfnamefont {U.}~\bibnamefont {Steiner}},\ and\
  \bibinfo {author} {\bibfnamefont {A.}~\bibnamefont {Sepe}},\ }\bibfield
  {title} {\bibinfo {title} {Synchrotron big data science},\ }\href
  {https://doi.org/10.1002/smll.201802291} {\bibfield  {journal} {\bibinfo
  {journal} {Small}\ }\textbf {\bibinfo {volume} {14}},\ \bibinfo {pages}
  {1802291} (\bibinfo {year} {2018})}\BibitemShut {NoStop}%
\bibitem [{\citenamefont {Greco}\ \emph {et~al.}(2019)\citenamefont {Greco},
  \citenamefont {Starostin}, \citenamefont {Karapanagiotis}, \citenamefont
  {Hinderhofer}, \citenamefont {Gerlach}, \citenamefont {Pithan}, \citenamefont
  {Liehr}, \citenamefont {Schreiber},\ and\ \citenamefont
  {Kowarik}}]{Greco_2019_JApplCrystallogr}%
  \BibitemOpen
  \bibfield  {author} {\bibinfo {author} {\bibfnamefont {A.}~\bibnamefont
  {Greco}}, \bibinfo {author} {\bibfnamefont {V.}~\bibnamefont {Starostin}},
  \bibinfo {author} {\bibfnamefont {C.}~\bibnamefont {Karapanagiotis}},
  \bibinfo {author} {\bibfnamefont {A.}~\bibnamefont {Hinderhofer}}, \bibinfo
  {author} {\bibfnamefont {A.}~\bibnamefont {Gerlach}}, \bibinfo {author}
  {\bibfnamefont {L.}~\bibnamefont {Pithan}}, \bibinfo {author} {\bibfnamefont
  {S.}~\bibnamefont {Liehr}}, \bibinfo {author} {\bibfnamefont
  {F.}~\bibnamefont {Schreiber}},\ and\ \bibinfo {author} {\bibfnamefont
  {S.}~\bibnamefont {Kowarik}},\ }\bibfield  {title} {\bibinfo {title} {Fast
  fitting of reflectivity data of growing thin films using neural networks},\
  }\href {https://doi.org/10.1107/S1600576719013311} {\bibfield  {journal}
  {\bibinfo  {journal} {J. Appl. Crystallogr.}\ }\textbf {\bibinfo {volume}
  {52}},\ \bibinfo {pages} {1342} (\bibinfo {year} {2019})}\BibitemShut
  {NoStop}%
\bibitem [{\citenamefont {Greco}\ \emph {et~al.}(2021)\citenamefont {Greco},
  \citenamefont {Starostin}, \citenamefont {Hinderhofer}, \citenamefont
  {Gerlach}, \citenamefont {Skoda}, \citenamefont {Kowarik},\ and\
  \citenamefont {Schreiber}}]{Greco2021}%
  \BibitemOpen
  \bibfield  {author} {\bibinfo {author} {\bibfnamefont {A.}~\bibnamefont
  {Greco}}, \bibinfo {author} {\bibfnamefont {V.}~\bibnamefont {Starostin}},
  \bibinfo {author} {\bibfnamefont {A.}~\bibnamefont {Hinderhofer}}, \bibinfo
  {author} {\bibfnamefont {A.}~\bibnamefont {Gerlach}}, \bibinfo {author}
  {\bibfnamefont {M.~W.~A.}\ \bibnamefont {Skoda}}, \bibinfo {author}
  {\bibfnamefont {S.}~\bibnamefont {Kowarik}},\ and\ \bibinfo {author}
  {\bibfnamefont {F.}~\bibnamefont {Schreiber}},\ }\bibfield  {title} {\bibinfo
  {title} {Neural network analysis of neutron and x-ray reflectivity data:
  pathological cases, performance and perspectives},\ }\href
  {https://doi.org/10.1088/2632-2153/abf9b1} {\bibfield  {journal} {\bibinfo
  {journal} {Machine Learning: Science and Technology}\ }\textbf {\bibinfo
  {volume} {2}},\ \bibinfo {pages} {045003} (\bibinfo {year}
  {2021})}\BibitemShut {NoStop}%
\bibitem [{\citenamefont {Samarakoon}\ \emph {et~al.}(2020)\citenamefont
  {Samarakoon}, \citenamefont {Barros}, \citenamefont {Li}, \citenamefont
  {Eisenbach}, \citenamefont {Zhang}, \citenamefont {Ye}, \citenamefont
  {Sharma}, \citenamefont {Dun}, \citenamefont {Zhou}, \citenamefont {Grigera},
  \citenamefont {Batista},\ and\ \citenamefont {Tennant}}]{Samarakoon2020}%
  \BibitemOpen
  \bibfield  {author} {\bibinfo {author} {\bibfnamefont {A.~M.}\ \bibnamefont
  {Samarakoon}}, \bibinfo {author} {\bibfnamefont {K.}~\bibnamefont {Barros}},
  \bibinfo {author} {\bibfnamefont {Y.~W.}\ \bibnamefont {Li}}, \bibinfo
  {author} {\bibfnamefont {M.}~\bibnamefont {Eisenbach}}, \bibinfo {author}
  {\bibfnamefont {Q.}~\bibnamefont {Zhang}}, \bibinfo {author} {\bibfnamefont
  {F.}~\bibnamefont {Ye}}, \bibinfo {author} {\bibfnamefont {V.}~\bibnamefont
  {Sharma}}, \bibinfo {author} {\bibfnamefont {Z.~L.}\ \bibnamefont {Dun}},
  \bibinfo {author} {\bibfnamefont {H.}~\bibnamefont {Zhou}}, \bibinfo {author}
  {\bibfnamefont {S.~A.}\ \bibnamefont {Grigera}}, \bibinfo {author}
  {\bibfnamefont {C.~D.}\ \bibnamefont {Batista}},\ and\ \bibinfo {author}
  {\bibfnamefont {D.~A.}\ \bibnamefont {Tennant}},\ }\bibfield  {title}
  {\bibinfo {title} {Machine-learning-assisted insight into spin ice
  dy2ti2o7},\ }\bibfield  {journal} {\bibinfo  {journal} {Nature
  Communications}\ }\textbf {\bibinfo {volume} {11}},\ \href
  {https://doi.org/10.1038/s41467-020-14660-y} {10.1038/s41467-020-14660-y}
  (\bibinfo {year} {2020})\BibitemShut {NoStop}%
\bibitem [{\citenamefont {Sanchez-Gonzalez}\ \emph {et~al.}(2017)\citenamefont
  {Sanchez-Gonzalez}, \citenamefont {Micaelli}, \citenamefont {Olivier},
  \citenamefont {Barillot}, \citenamefont {Ilchen}, \citenamefont {Lutman},
  \citenamefont {Marinelli}, \citenamefont {Maxwell}, \citenamefont {Achner},
  \citenamefont {Ag{\aa}ker}, \citenamefont {Berrah}, \citenamefont {Bostedt},
  \citenamefont {Bozek}, \citenamefont {Buck}, \citenamefont {Bucksbaum},
  \citenamefont {Montero}, \citenamefont {Cooper}, \citenamefont {Cryan},
  \citenamefont {Dong}, \citenamefont {Feifel}, \citenamefont {Frasinski},
  \citenamefont {Fukuzawa}, \citenamefont {Galler}, \citenamefont {Hartmann},
  \citenamefont {Hartmann}, \citenamefont {Helml}, \citenamefont {Johnson},
  \citenamefont {Knie}, \citenamefont {Lindahl}, \citenamefont {Liu},
  \citenamefont {Motomura}, \citenamefont {Mucke}, \citenamefont {O'Grady},
  \citenamefont {Rubensson}, \citenamefont {Simpson}, \citenamefont {Squibb},
  \citenamefont {S{\aa}the}, \citenamefont {Ueda}, \citenamefont {Vacher},
  \citenamefont {Walke}, \citenamefont {Zhaunerchyk}, \citenamefont {Coffee},\
  and\ \citenamefont {Marangos}}]{SanchezGonzalez2017}%
  \BibitemOpen
  \bibfield  {author} {\bibinfo {author} {\bibfnamefont {A.}~\bibnamefont
  {Sanchez-Gonzalez}}, \bibinfo {author} {\bibfnamefont {P.}~\bibnamefont
  {Micaelli}}, \bibinfo {author} {\bibfnamefont {C.}~\bibnamefont {Olivier}},
  \bibinfo {author} {\bibfnamefont {T.~R.}\ \bibnamefont {Barillot}}, \bibinfo
  {author} {\bibfnamefont {M.}~\bibnamefont {Ilchen}}, \bibinfo {author}
  {\bibfnamefont {A.~A.}\ \bibnamefont {Lutman}}, \bibinfo {author}
  {\bibfnamefont {A.}~\bibnamefont {Marinelli}}, \bibinfo {author}
  {\bibfnamefont {T.}~\bibnamefont {Maxwell}}, \bibinfo {author} {\bibfnamefont
  {A.}~\bibnamefont {Achner}}, \bibinfo {author} {\bibfnamefont
  {M.}~\bibnamefont {Ag{\aa}ker}}, \bibinfo {author} {\bibfnamefont
  {N.}~\bibnamefont {Berrah}}, \bibinfo {author} {\bibfnamefont
  {C.}~\bibnamefont {Bostedt}}, \bibinfo {author} {\bibfnamefont {J.~D.}\
  \bibnamefont {Bozek}}, \bibinfo {author} {\bibfnamefont {J.}~\bibnamefont
  {Buck}}, \bibinfo {author} {\bibfnamefont {P.~H.}\ \bibnamefont {Bucksbaum}},
  \bibinfo {author} {\bibfnamefont {S.~C.}\ \bibnamefont {Montero}}, \bibinfo
  {author} {\bibfnamefont {B.}~\bibnamefont {Cooper}}, \bibinfo {author}
  {\bibfnamefont {J.~P.}\ \bibnamefont {Cryan}}, \bibinfo {author}
  {\bibfnamefont {M.}~\bibnamefont {Dong}}, \bibinfo {author} {\bibfnamefont
  {R.}~\bibnamefont {Feifel}}, \bibinfo {author} {\bibfnamefont {L.~J.}\
  \bibnamefont {Frasinski}}, \bibinfo {author} {\bibfnamefont {H.}~\bibnamefont
  {Fukuzawa}}, \bibinfo {author} {\bibfnamefont {A.}~\bibnamefont {Galler}},
  \bibinfo {author} {\bibfnamefont {G.}~\bibnamefont {Hartmann}}, \bibinfo
  {author} {\bibfnamefont {N.}~\bibnamefont {Hartmann}}, \bibinfo {author}
  {\bibfnamefont {W.}~\bibnamefont {Helml}}, \bibinfo {author} {\bibfnamefont
  {A.~S.}\ \bibnamefont {Johnson}}, \bibinfo {author} {\bibfnamefont
  {A.}~\bibnamefont {Knie}}, \bibinfo {author} {\bibfnamefont {A.~O.}\
  \bibnamefont {Lindahl}}, \bibinfo {author} {\bibfnamefont {J.}~\bibnamefont
  {Liu}}, \bibinfo {author} {\bibfnamefont {K.}~\bibnamefont {Motomura}},
  \bibinfo {author} {\bibfnamefont {M.}~\bibnamefont {Mucke}}, \bibinfo
  {author} {\bibfnamefont {C.}~\bibnamefont {O'Grady}}, \bibinfo {author}
  {\bibfnamefont {J.-E.}\ \bibnamefont {Rubensson}}, \bibinfo {author}
  {\bibfnamefont {E.~R.}\ \bibnamefont {Simpson}}, \bibinfo {author}
  {\bibfnamefont {R.~J.}\ \bibnamefont {Squibb}}, \bibinfo {author}
  {\bibfnamefont {C.}~\bibnamefont {S{\aa}the}}, \bibinfo {author}
  {\bibfnamefont {K.}~\bibnamefont {Ueda}}, \bibinfo {author} {\bibfnamefont
  {M.}~\bibnamefont {Vacher}}, \bibinfo {author} {\bibfnamefont {D.~J.}\
  \bibnamefont {Walke}}, \bibinfo {author} {\bibfnamefont {V.}~\bibnamefont
  {Zhaunerchyk}}, \bibinfo {author} {\bibfnamefont {R.~N.}\ \bibnamefont
  {Coffee}},\ and\ \bibinfo {author} {\bibfnamefont {J.~P.}\ \bibnamefont
  {Marangos}},\ }\bibfield  {title} {\bibinfo {title} {Accurate prediction of
  x-ray pulse properties from a free-electron laser using machine learning},\
  }\bibfield  {journal} {\bibinfo  {journal} {Nature Communications}\ }\textbf
  {\bibinfo {volume} {8}},\ \href {https://doi.org/10.1038/ncomms15461}
  {10.1038/ncomms15461} (\bibinfo {year} {2017})\BibitemShut {NoStop}%
\bibitem [{\citenamefont {Ziletti}\ \emph {et~al.}(2018)\citenamefont
  {Ziletti}, \citenamefont {Kumar}, \citenamefont {Scheffler},\ and\
  \citenamefont {Ghiringhelli}}]{Ziletti2018}%
  \BibitemOpen
  \bibfield  {author} {\bibinfo {author} {\bibfnamefont {A.}~\bibnamefont
  {Ziletti}}, \bibinfo {author} {\bibfnamefont {D.}~\bibnamefont {Kumar}},
  \bibinfo {author} {\bibfnamefont {M.}~\bibnamefont {Scheffler}},\ and\
  \bibinfo {author} {\bibfnamefont {L.~M.}\ \bibnamefont {Ghiringhelli}},\
  }\bibfield  {title} {\bibinfo {title} {Insightful classification of crystal
  structures using deep learning},\ }\bibfield  {journal} {\bibinfo  {journal}
  {Nature Communications}\ }\textbf {\bibinfo {volume} {9}},\ \href
  {https://doi.org/10.1038/s41467-018-05169-6} {10.1038/s41467-018-05169-6}
  (\bibinfo {year} {2018})\BibitemShut {NoStop}%
\bibitem [{\citenamefont {Ryan}\ \emph {et~al.}(2018)\citenamefont {Ryan},
  \citenamefont {Lengyel},\ and\ \citenamefont {Shatruk}}]{Ryan2018}%
  \BibitemOpen
  \bibfield  {author} {\bibinfo {author} {\bibfnamefont {K.}~\bibnamefont
  {Ryan}}, \bibinfo {author} {\bibfnamefont {J.}~\bibnamefont {Lengyel}},\ and\
  \bibinfo {author} {\bibfnamefont {M.}~\bibnamefont {Shatruk}},\ }\bibfield
  {title} {\bibinfo {title} {Crystal structure prediction via deep learning},\
  }\href {https://doi.org/10.1021/jacs.8b03913} {\bibfield  {journal} {\bibinfo
   {journal} {Journal of the American Chemical Society}\ }\textbf {\bibinfo
  {volume} {140}},\ \bibinfo {pages} {10158} (\bibinfo {year}
  {2018})}\BibitemShut {NoStop}%
\bibitem [{\citenamefont {Oviedo}\ \emph {et~al.}(2019)\citenamefont {Oviedo},
  \citenamefont {Ren}, \citenamefont {Sun}, \citenamefont {Settens},
  \citenamefont {Liu}, \citenamefont {Hartono}, \citenamefont {Ramasamy},
  \citenamefont {DeCost}, \citenamefont {Tian}, \citenamefont {Romano},
  \citenamefont {Kusne},\ and\ \citenamefont {Buonassisi}}]{Oviedo2019}%
  \BibitemOpen
  \bibfield  {author} {\bibinfo {author} {\bibfnamefont {F.}~\bibnamefont
  {Oviedo}}, \bibinfo {author} {\bibfnamefont {Z.}~\bibnamefont {Ren}},
  \bibinfo {author} {\bibfnamefont {S.}~\bibnamefont {Sun}}, \bibinfo {author}
  {\bibfnamefont {C.}~\bibnamefont {Settens}}, \bibinfo {author} {\bibfnamefont
  {Z.}~\bibnamefont {Liu}}, \bibinfo {author} {\bibfnamefont {N.~T.~P.}\
  \bibnamefont {Hartono}}, \bibinfo {author} {\bibfnamefont {S.}~\bibnamefont
  {Ramasamy}}, \bibinfo {author} {\bibfnamefont {B.~L.}\ \bibnamefont
  {DeCost}}, \bibinfo {author} {\bibfnamefont {S.~I.~P.}\ \bibnamefont {Tian}},
  \bibinfo {author} {\bibfnamefont {G.}~\bibnamefont {Romano}}, \bibinfo
  {author} {\bibfnamefont {A.~G.}\ \bibnamefont {Kusne}},\ and\ \bibinfo
  {author} {\bibfnamefont {T.}~\bibnamefont {Buonassisi}},\ }\bibfield  {title}
  {\bibinfo {title} {Fast and interpretable classification of small x-ray
  diffraction datasets using data augmentation and deep neural networks},\
  }\bibfield  {journal} {\bibinfo  {journal} {npj Computational Materials}\
  }\textbf {\bibinfo {volume} {5}},\ \href
  {https://doi.org/10.1038/s41524-019-0196-x} {10.1038/s41524-019-0196-x}
  (\bibinfo {year} {2019})\BibitemShut {NoStop}%
\bibitem [{\citenamefont {Tatlier}(2011)}]{tatlier2011artificial}%
  \BibitemOpen
  \bibfield  {author} {\bibinfo {author} {\bibfnamefont {M.}~\bibnamefont
  {Tatlier}},\ }\bibfield  {title} {\bibinfo {title} {Artificial neural network
  methods for the prediction of framework crystal structures of zeolites from
  xrd data},\ }\href@noop {} {\bibfield  {journal} {\bibinfo  {journal} {Neural
  Computing and Applications}\ }\textbf {\bibinfo {volume} {20}},\ \bibinfo
  {pages} {365} (\bibinfo {year} {2011})}\BibitemShut {NoStop}%
\bibitem [{\citenamefont {Lee}\ \emph {et~al.}(2020)\citenamefont {Lee},
  \citenamefont {Park}, \citenamefont {Lee}, \citenamefont {Singh},\ and\
  \citenamefont {Sohn}}]{Lee2020}%
  \BibitemOpen
  \bibfield  {author} {\bibinfo {author} {\bibfnamefont {J.-W.}\ \bibnamefont
  {Lee}}, \bibinfo {author} {\bibfnamefont {W.~B.}\ \bibnamefont {Park}},
  \bibinfo {author} {\bibfnamefont {J.~H.}\ \bibnamefont {Lee}}, \bibinfo
  {author} {\bibfnamefont {S.~P.}\ \bibnamefont {Singh}},\ and\ \bibinfo
  {author} {\bibfnamefont {K.-S.}\ \bibnamefont {Sohn}},\ }\bibfield  {title}
  {\bibinfo {title} {A deep-learning technique for phase identification in
  multiphase inorganic compounds using synthetic {XRD} powder patterns},\
  }\bibfield  {journal} {\bibinfo  {journal} {Nature Communications}\ }\textbf
  {\bibinfo {volume} {11}},\ \href {https://doi.org/10.1038/s41467-019-13749-3}
  {10.1038/s41467-019-13749-3} (\bibinfo {year} {2020})\BibitemShut {NoStop}%
\bibitem [{\citenamefont {Wang}\ \emph {et~al.}(2017)\citenamefont {Wang},
  \citenamefont {Yager}, \citenamefont {Yu},\ and\ \citenamefont
  {Hoai}}]{wang2017x}%
  \BibitemOpen
  \bibfield  {author} {\bibinfo {author} {\bibfnamefont {B.}~\bibnamefont
  {Wang}}, \bibinfo {author} {\bibfnamefont {K.}~\bibnamefont {Yager}},
  \bibinfo {author} {\bibfnamefont {D.}~\bibnamefont {Yu}},\ and\ \bibinfo
  {author} {\bibfnamefont {M.}~\bibnamefont {Hoai}},\ }\bibfield  {title}
  {\bibinfo {title} {X-ray scattering image classification using deep
  learning},\ }in\ \href@noop {} {\emph {\bibinfo {booktitle} {2017 IEEE Winter
  Conference on Applications of Computer Vision (WACV)}}}\ (\bibinfo
  {organization} {IEEE},\ \bibinfo {year} {2017})\ pp.\ \bibinfo {pages}
  {697--704}\BibitemShut {NoStop}%
\bibitem [{\citenamefont {Ke}\ \emph {et~al.}(2018)\citenamefont {Ke},
  \citenamefont {Brewster}, \citenamefont {Yu}, \citenamefont {Ushizima},
  \citenamefont {Yang},\ and\ \citenamefont {Sauter}}]{ke2018convolutional}%
  \BibitemOpen
  \bibfield  {author} {\bibinfo {author} {\bibfnamefont {T.-W.}\ \bibnamefont
  {Ke}}, \bibinfo {author} {\bibfnamefont {A.~S.}\ \bibnamefont {Brewster}},
  \bibinfo {author} {\bibfnamefont {S.~X.}\ \bibnamefont {Yu}}, \bibinfo
  {author} {\bibfnamefont {D.}~\bibnamefont {Ushizima}}, \bibinfo {author}
  {\bibfnamefont {C.}~\bibnamefont {Yang}},\ and\ \bibinfo {author}
  {\bibfnamefont {N.~K.}\ \bibnamefont {Sauter}},\ }\bibfield  {title}
  {\bibinfo {title} {A convolutional neural network-based screening tool for
  x-ray serial crystallography},\ }\href@noop {} {\bibfield  {journal}
  {\bibinfo  {journal} {Journal of synchrotron radiation}\ }\textbf {\bibinfo
  {volume} {25}},\ \bibinfo {pages} {655} (\bibinfo {year} {2018})}\BibitemShut
  {NoStop}%
\bibitem [{\citenamefont {Liu}\ \emph {et~al.}(2019)\citenamefont {Liu},
  \citenamefont {Melton}, \citenamefont {Venkatakrishnan}, \citenamefont
  {Pandolfi}, \citenamefont {Freychet}, \citenamefont {Kumar}, \citenamefont
  {Tang}, \citenamefont {Hexemer},\ and\ \citenamefont
  {Ushizima}}]{Liu_2019_MRSCommunications}%
  \BibitemOpen
  \bibfield  {author} {\bibinfo {author} {\bibfnamefont {S.}~\bibnamefont
  {Liu}}, \bibinfo {author} {\bibfnamefont {C.~N.}\ \bibnamefont {Melton}},
  \bibinfo {author} {\bibfnamefont {S.}~\bibnamefont {Venkatakrishnan}},
  \bibinfo {author} {\bibfnamefont {R.~J.}\ \bibnamefont {Pandolfi}}, \bibinfo
  {author} {\bibfnamefont {G.}~\bibnamefont {Freychet}}, \bibinfo {author}
  {\bibfnamefont {D.}~\bibnamefont {Kumar}}, \bibinfo {author} {\bibfnamefont
  {H.}~\bibnamefont {Tang}}, \bibinfo {author} {\bibfnamefont {A.}~\bibnamefont
  {Hexemer}},\ and\ \bibinfo {author} {\bibfnamefont {D.~M.}\ \bibnamefont
  {Ushizima}},\ }\bibfield  {title} {\bibinfo {title} {Convolutional neural
  networks for grazing incidence x-ray scattering patterns: thin film structure
  identification},\ }\href {https://doi.org/10.1557/mrc.2019.26} {\bibfield
  {journal} {\bibinfo  {journal} {MRS Communications}\ }\textbf {\bibinfo
  {volume} {9}},\ \bibinfo {pages} {586} (\bibinfo {year} {2019})}\BibitemShut
  {NoStop}%
\bibitem [{\citenamefont {Sullivan}\ \emph {et~al.}(2019)\citenamefont
  {Sullivan}, \citenamefont {Archibald}, \citenamefont {Azadmanesh},
  \citenamefont {Vandavasi}, \citenamefont {Langan}, \citenamefont {Coates},
  \citenamefont {Lynch},\ and\ \citenamefont {Langan}}]{BraggNet2019}%
  \BibitemOpen
  \bibfield  {author} {\bibinfo {author} {\bibfnamefont {B.}~\bibnamefont
  {Sullivan}}, \bibinfo {author} {\bibfnamefont {R.}~\bibnamefont {Archibald}},
  \bibinfo {author} {\bibfnamefont {J.}~\bibnamefont {Azadmanesh}}, \bibinfo
  {author} {\bibfnamefont {V.~G.}\ \bibnamefont {Vandavasi}}, \bibinfo {author}
  {\bibfnamefont {P.~S.}\ \bibnamefont {Langan}}, \bibinfo {author}
  {\bibfnamefont {L.}~\bibnamefont {Coates}}, \bibinfo {author} {\bibfnamefont
  {V.}~\bibnamefont {Lynch}},\ and\ \bibinfo {author} {\bibfnamefont
  {P.}~\bibnamefont {Langan}},\ }\bibfield  {title} {\bibinfo {title}
  {{BraggNet}: integrating bragg peaks using neural networks},\ }\href
  {https://doi.org/10.1107/s1600576719008665} {\bibfield  {journal} {\bibinfo
  {journal} {Journal of Applied Crystallography}\ }\textbf {\bibinfo {volume}
  {52}},\ \bibinfo {pages} {854} (\bibinfo {year} {2019})}\BibitemShut
  {NoStop}%
\bibitem [{\citenamefont {Liu}\ \emph {et~al.}(2020{\natexlab{b}})\citenamefont
  {Liu}, \citenamefont {Sharma}, \citenamefont {Park}, \citenamefont {Kenesei},
  \citenamefont {Almer}, \citenamefont {Kettimuthu},\ and\ \citenamefont
  {Foster}}]{braggnn2020}%
  \BibitemOpen
  \bibfield  {author} {\bibinfo {author} {\bibfnamefont {Z.}~\bibnamefont
  {Liu}}, \bibinfo {author} {\bibfnamefont {H.}~\bibnamefont {Sharma}},
  \bibinfo {author} {\bibfnamefont {J.-S.}\ \bibnamefont {Park}}, \bibinfo
  {author} {\bibfnamefont {P.}~\bibnamefont {Kenesei}}, \bibinfo {author}
  {\bibfnamefont {J.}~\bibnamefont {Almer}}, \bibinfo {author} {\bibfnamefont
  {R.}~\bibnamefont {Kettimuthu}},\ and\ \bibinfo {author} {\bibfnamefont
  {I.}~\bibnamefont {Foster}},\ }\bibfield  {title} {\bibinfo {title} {Braggnn:
  Fast x-ray bragg peak analysis using deep learning},\ }\href@noop {}
  {\bibfield  {journal} {\bibinfo  {journal} {arXiv preprint arXiv:2008.08198}\
  } (\bibinfo {year} {2020}{\natexlab{b}})}\BibitemShut {NoStop}%
\bibitem [{\citenamefont {Ren}\ \emph {et~al.}(2015)\citenamefont {Ren},
  \citenamefont {He}, \citenamefont {Girshick},\ and\ \citenamefont
  {Sun}}]{fasterrcnn2015}%
  \BibitemOpen
  \bibfield  {author} {\bibinfo {author} {\bibfnamefont {S.}~\bibnamefont
  {Ren}}, \bibinfo {author} {\bibfnamefont {K.}~\bibnamefont {He}}, \bibinfo
  {author} {\bibfnamefont {R.}~\bibnamefont {Girshick}},\ and\ \bibinfo
  {author} {\bibfnamefont {J.}~\bibnamefont {Sun}},\ }\bibfield  {title}
  {\bibinfo {title} {Faster {R-CNN}: Towards real-time object detection with
  region proposal networks},\ }in\ \href
  {https://proceedings.neurips.cc/paper/2015/file/14bfa6bb14875e45bba028a21ed38046-Paper.pdf}
  {\emph {\bibinfo {booktitle} {Advances in Neural Information Processing
  Systems}}},\ Vol.~\bibinfo {volume} {28},\ \bibinfo {editor} {edited by\
  \bibinfo {editor} {\bibfnamefont {C.}~\bibnamefont {Cortes}}, \bibinfo
  {editor} {\bibfnamefont {N.}~\bibnamefont {Lawrence}}, \bibinfo {editor}
  {\bibfnamefont {D.}~\bibnamefont {Lee}}, \bibinfo {editor} {\bibfnamefont
  {M.}~\bibnamefont {Sugiyama}},\ and\ \bibinfo {editor} {\bibfnamefont
  {R.}~\bibnamefont {Garnett}}}\ (\bibinfo  {publisher} {Curran Associates,
  Inc.},\ \bibinfo {year} {2015})\BibitemShut {NoStop}%
\bibitem [{\citenamefont {Lin}\ \emph {et~al.}(2017)\citenamefont {Lin},
  \citenamefont {Dollar}, \citenamefont {Girshick}, \citenamefont {He},
  \citenamefont {Hariharan},\ and\ \citenamefont
  {Belongie}}]{feature_pyramid_networks_2017}%
  \BibitemOpen
  \bibfield  {author} {\bibinfo {author} {\bibfnamefont {T.-Y.}\ \bibnamefont
  {Lin}}, \bibinfo {author} {\bibfnamefont {P.}~\bibnamefont {Dollar}},
  \bibinfo {author} {\bibfnamefont {R.}~\bibnamefont {Girshick}}, \bibinfo
  {author} {\bibfnamefont {K.}~\bibnamefont {He}}, \bibinfo {author}
  {\bibfnamefont {B.}~\bibnamefont {Hariharan}},\ and\ \bibinfo {author}
  {\bibfnamefont {S.}~\bibnamefont {Belongie}},\ }\bibfield  {title} {\bibinfo
  {title} {Feature pyramid networks for object detection},\ }in\ \href
  {https://doi.org/10.1109/cvpr.2017.106} {\emph {\bibinfo {booktitle} {2017
  {IEEE} Conference on Computer Vision and Pattern Recognition ({CVPR})}}}\
  (\bibinfo  {publisher} {{IEEE}},\ \bibinfo {year} {2017})\BibitemShut
  {NoStop}%
\bibitem [{\citenamefont {Girshick}(2015)}]{fastrcnn2015}%
  \BibitemOpen
  \bibfield  {author} {\bibinfo {author} {\bibfnamefont {R.}~\bibnamefont
  {Girshick}},\ }\bibfield  {title} {\bibinfo {title} {Fast {R-CNN}},\ }in\
  \href@noop {} {\emph {\bibinfo {booktitle} {Proceedings of the IEEE
  international conference on computer vision}}}\ (\bibinfo {year} {2015})\
  pp.\ \bibinfo {pages} {1440--1448}\BibitemShut {NoStop}%
\bibitem [{\citenamefont {He}\ \emph {et~al.}(2017)\citenamefont {He},
  \citenamefont {Gkioxari}, \citenamefont {Doll{\'a}r},\ and\ \citenamefont
  {Girshick}}]{maskrcnn2017}%
  \BibitemOpen
  \bibfield  {author} {\bibinfo {author} {\bibfnamefont {K.}~\bibnamefont
  {He}}, \bibinfo {author} {\bibfnamefont {G.}~\bibnamefont {Gkioxari}},
  \bibinfo {author} {\bibfnamefont {P.}~\bibnamefont {Doll{\'a}r}},\ and\
  \bibinfo {author} {\bibfnamefont {R.}~\bibnamefont {Girshick}},\ }\bibfield
  {title} {\bibinfo {title} {Mask {R-CNN}},\ }in\ \href@noop {} {\emph
  {\bibinfo {booktitle} {Proceedings of the IEEE international conference on
  computer vision}}}\ (\bibinfo {year} {2017})\ pp.\ \bibinfo {pages}
  {2961--2969}\BibitemShut {NoStop}%
\bibitem [{\citenamefont {Dai}\ \emph {et~al.}(2015)\citenamefont {Dai},
  \citenamefont {He},\ and\ \citenamefont {Sun}}]{dai2015instanceaware}%
  \BibitemOpen
  \bibfield  {author} {\bibinfo {author} {\bibfnamefont {J.}~\bibnamefont
  {Dai}}, \bibinfo {author} {\bibfnamefont {K.}~\bibnamefont {He}},\ and\
  \bibinfo {author} {\bibfnamefont {J.}~\bibnamefont {Sun}},\ }\href@noop {}
  {\bibinfo {title} {Instance-aware semantic segmentation via multi-task
  network cascades}} (\bibinfo {year} {2015}),\ \Eprint
  {https://arxiv.org/abs/1512.04412} {arXiv:1512.04412 [cs.CV]} \BibitemShut
  {NoStop}%
\bibitem [{\citenamefont {Redmon}\ \emph {et~al.}(2016)\citenamefont {Redmon},
  \citenamefont {Divvala}, \citenamefont {Girshick},\ and\ \citenamefont
  {Farhadi}}]{yolo2016}%
  \BibitemOpen
  \bibfield  {author} {\bibinfo {author} {\bibfnamefont {J.}~\bibnamefont
  {Redmon}}, \bibinfo {author} {\bibfnamefont {S.}~\bibnamefont {Divvala}},
  \bibinfo {author} {\bibfnamefont {R.}~\bibnamefont {Girshick}},\ and\
  \bibinfo {author} {\bibfnamefont {A.}~\bibnamefont {Farhadi}},\ }\bibfield
  {title} {\bibinfo {title} {You only look once: Unified, real-time object
  detection},\ }in\ \href@noop {} {\emph {\bibinfo {booktitle} {Proceedings of
  the IEEE conference on computer vision and pattern recognition}}}\ (\bibinfo
  {year} {2016})\ pp.\ \bibinfo {pages} {779--788}\BibitemShut {NoStop}%
\bibitem [{\citenamefont {Carion}\ \emph {et~al.}(2020)\citenamefont {Carion},
  \citenamefont {Massa}, \citenamefont {Synnaeve}, \citenamefont {Usunier},
  \citenamefont {Kirillov},\ and\ \citenamefont
  {Zagoruyko}}]{transformer_detr_2020}%
  \BibitemOpen
  \bibfield  {author} {\bibinfo {author} {\bibfnamefont {N.}~\bibnamefont
  {Carion}}, \bibinfo {author} {\bibfnamefont {F.}~\bibnamefont {Massa}},
  \bibinfo {author} {\bibfnamefont {G.}~\bibnamefont {Synnaeve}}, \bibinfo
  {author} {\bibfnamefont {N.}~\bibnamefont {Usunier}}, \bibinfo {author}
  {\bibfnamefont {A.}~\bibnamefont {Kirillov}},\ and\ \bibinfo {author}
  {\bibfnamefont {S.}~\bibnamefont {Zagoruyko}},\ }\bibfield  {title} {\bibinfo
  {title} {End-to-end object detection with transformers},\ }in\ \href@noop {}
  {\emph {\bibinfo {booktitle} {European Conference on Computer Vision}}}\
  (\bibinfo {organization} {Springer},\ \bibinfo {year} {2020})\ pp.\ \bibinfo
  {pages} {213--229}\BibitemShut {NoStop}%
\bibitem [{\citenamefont {Rietveld}(1967)}]{Rietveld1967}%
  \BibitemOpen
  \bibfield  {author} {\bibinfo {author} {\bibfnamefont {H.~M.}\ \bibnamefont
  {Rietveld}},\ }\bibfield  {title} {\bibinfo {title} {Line profiles of neutron
  powder-diffraction peaks for structure refinement},\ }\href
  {https://doi.org/10.1107/s0365110x67000234} {\bibfield  {journal} {\bibinfo
  {journal} {Acta Crystallographica}\ }\textbf {\bibinfo {volume} {22}},\
  \bibinfo {pages} {151} (\bibinfo {year} {1967})}\BibitemShut {NoStop}%
\bibitem [{\citenamefont {Rietveld}(1969)}]{Rietveld1969}%
  \BibitemOpen
  \bibfield  {author} {\bibinfo {author} {\bibfnamefont {H.~M.}\ \bibnamefont
  {Rietveld}},\ }\bibfield  {title} {\bibinfo {title} {A profile refinement
  method for nuclear and magnetic structures},\ }\href
  {https://doi.org/10.1107/s0021889869006558} {\bibfield  {journal} {\bibinfo
  {journal} {J. Appl. Crystallogr.}\ }\textbf {\bibinfo {volume} {2}},\
  \bibinfo {pages} {65} (\bibinfo {year} {1969})}\BibitemShut {NoStop}%
\bibitem [{\citenamefont {David}(1986)}]{David1986}%
  \BibitemOpen
  \bibfield  {author} {\bibinfo {author} {\bibfnamefont {W.~I.~F.}\
  \bibnamefont {David}},\ }\bibfield  {title} {\bibinfo {title} {Powder
  diffraction peak shapes. parameterization of the pseudo-voigt as a voigt
  function},\ }\href {https://doi.org/10.1107/s0021889886089999} {\bibfield
  {journal} {\bibinfo  {journal} {J. Appl. Crystallogr.}\ }\textbf {\bibinfo
  {volume} {19}},\ \bibinfo {pages} {63} (\bibinfo {year} {1986})}\BibitemShut
  {NoStop}%
\bibitem [{\citenamefont {Dinapoli}\ \emph {et~al.}(2011)\citenamefont
  {Dinapoli}, \citenamefont {Bergamaschi}, \citenamefont {Henrich},
  \citenamefont {Horisberger}, \citenamefont {Johnson}, \citenamefont
  {Mozzanica}, \citenamefont {Schmid}, \citenamefont {Schmitt}, \citenamefont
  {Schreiber}, \citenamefont {Shi},\ and\ \citenamefont {Theidel}}]{EIGER}%
  \BibitemOpen
  \bibfield  {author} {\bibinfo {author} {\bibfnamefont {R.}~\bibnamefont
  {Dinapoli}}, \bibinfo {author} {\bibfnamefont {A.}~\bibnamefont
  {Bergamaschi}}, \bibinfo {author} {\bibfnamefont {B.}~\bibnamefont
  {Henrich}}, \bibinfo {author} {\bibfnamefont {R.}~\bibnamefont
  {Horisberger}}, \bibinfo {author} {\bibfnamefont {I.}~\bibnamefont
  {Johnson}}, \bibinfo {author} {\bibfnamefont {A.}~\bibnamefont {Mozzanica}},
  \bibinfo {author} {\bibfnamefont {E.}~\bibnamefont {Schmid}}, \bibinfo
  {author} {\bibfnamefont {B.}~\bibnamefont {Schmitt}}, \bibinfo {author}
  {\bibfnamefont {A.}~\bibnamefont {Schreiber}}, \bibinfo {author}
  {\bibfnamefont {X.}~\bibnamefont {Shi}},\ and\ \bibinfo {author}
  {\bibfnamefont {G.}~\bibnamefont {Theidel}},\ }\bibfield  {title} {\bibinfo
  {title} {Eiger: Next generation single photon counting detector for x-ray
  applications},\ }\href
  {https://doi.org/https://doi.org/10.1016/j.nima.2010.12.005} {\bibfield
  {journal} {\bibinfo  {journal} {Nuclear Instruments and Methods in Physics
  Research Section A: Accelerators, Spectrometers, Detectors and Associated
  Equipment}\ }\textbf {\bibinfo {volume} {650}},\ \bibinfo {pages} {79}
  (\bibinfo {year} {2011})}\BibitemShut {NoStop}%
\bibitem [{\citenamefont {Ronneberger}\ \emph {et~al.}(2015)\citenamefont
  {Ronneberger}, \citenamefont {Fischer},\ and\ \citenamefont
  {Brox}}]{unet_2015}%
  \BibitemOpen
  \bibfield  {author} {\bibinfo {author} {\bibfnamefont {O.}~\bibnamefont
  {Ronneberger}}, \bibinfo {author} {\bibfnamefont {P.}~\bibnamefont
  {Fischer}},\ and\ \bibinfo {author} {\bibfnamefont {T.}~\bibnamefont
  {Brox}},\ }\bibfield  {title} {\bibinfo {title} {U-net: Convolutional
  networks for biomedical image segmentation},\ }in\ \href
  {https://doi.org/10.1007/978-3-319-24574-4_28} {\emph {\bibinfo {booktitle}
  {Lecture Notes in Computer Science}}}\ (\bibinfo  {publisher} {Springer
  International Publishing},\ \bibinfo {year} {2015})\ pp.\ \bibinfo {pages}
  {234--241}\BibitemShut {NoStop}%
\bibitem [{\citenamefont {Wang}\ \emph {et~al.}(2021)\citenamefont {Wang},
  \citenamefont {Yeh},\ and\ \citenamefont {Liao}}]{wang2021you}%
  \BibitemOpen
  \bibfield  {author} {\bibinfo {author} {\bibfnamefont {C.-Y.}\ \bibnamefont
  {Wang}}, \bibinfo {author} {\bibfnamefont {I.-H.}\ \bibnamefont {Yeh}},\ and\
  \bibinfo {author} {\bibfnamefont {H.-Y.~M.}\ \bibnamefont {Liao}},\
  }\bibfield  {title} {\bibinfo {title} {You only learn one representation:
  Unified network for multiple tasks},\ }\href@noop {} {\bibfield  {journal}
  {\bibinfo  {journal} {arXiv preprint arXiv:2105.04206}\ } (\bibinfo {year}
  {2021})}\BibitemShut {NoStop}%
\bibitem [{\citenamefont {Stoumpos}\ \emph {et~al.}(2013)\citenamefont
  {Stoumpos}, \citenamefont {Malliakas},\ and\ \citenamefont
  {Kanatzidis}}]{Stoumpos2013}%
  \BibitemOpen
  \bibfield  {author} {\bibinfo {author} {\bibfnamefont {C.~C.}\ \bibnamefont
  {Stoumpos}}, \bibinfo {author} {\bibfnamefont {C.~D.}\ \bibnamefont
  {Malliakas}},\ and\ \bibinfo {author} {\bibfnamefont {M.~G.}\ \bibnamefont
  {Kanatzidis}},\ }\bibfield  {title} {\bibinfo {title} {Semiconducting tin and
  lead iodide perovskites with organic cations: Phase transitions, high
  mobilities, and near-infrared photoluminescent properties},\ }\href
  {https://doi.org/10.1021/ic401215x} {\bibfield  {journal} {\bibinfo
  {journal} {Inorganic Chemistry}\ }\textbf {\bibinfo {volume} {52}},\ \bibinfo
  {pages} {9019} (\bibinfo {year} {2013})}\BibitemShut {NoStop}%
\bibitem [{\citenamefont {Stoumpos}\ \emph {et~al.}(2016)\citenamefont
  {Stoumpos}, \citenamefont {Cao}, \citenamefont {Clark}, \citenamefont
  {Young}, \citenamefont {Rondinelli}, \citenamefont {Jang}, \citenamefont
  {Hupp},\ and\ \citenamefont {Kanatzidis}}]{Stoumpos2016}%
  \BibitemOpen
  \bibfield  {author} {\bibinfo {author} {\bibfnamefont {C.~C.}\ \bibnamefont
  {Stoumpos}}, \bibinfo {author} {\bibfnamefont {D.~H.}\ \bibnamefont {Cao}},
  \bibinfo {author} {\bibfnamefont {D.~J.}\ \bibnamefont {Clark}}, \bibinfo
  {author} {\bibfnamefont {J.}~\bibnamefont {Young}}, \bibinfo {author}
  {\bibfnamefont {J.~M.}\ \bibnamefont {Rondinelli}}, \bibinfo {author}
  {\bibfnamefont {J.~I.}\ \bibnamefont {Jang}}, \bibinfo {author}
  {\bibfnamefont {J.~T.}\ \bibnamefont {Hupp}},\ and\ \bibinfo {author}
  {\bibfnamefont {M.~G.}\ \bibnamefont {Kanatzidis}},\ }\bibfield  {title}
  {\bibinfo {title} {Ruddlesden{\textendash}popper hybrid lead iodide
  perovskite 2d homologous semiconductors},\ }\href
  {https://doi.org/10.1021/acs.chemmater.6b00847} {\bibfield  {journal}
  {\bibinfo  {journal} {Chemistry of Materials}\ }\textbf {\bibinfo {volume}
  {28}},\ \bibinfo {pages} {2852} (\bibinfo {year} {2016})}\BibitemShut
  {NoStop}%
\bibitem [{\citenamefont {Soe}\ \emph {et~al.}(2018)\citenamefont {Soe},
  \citenamefont {Nagabhushana}, \citenamefont {Shivaramaiah}, \citenamefont
  {Tsai}, \citenamefont {Nie}, \citenamefont {Blancon}, \citenamefont
  {Melkonyan}, \citenamefont {Cao}, \citenamefont {Traor{\'{e}}}, \citenamefont
  {Pedesseau}, \citenamefont {Kepenekian}, \citenamefont {Katan}, \citenamefont
  {Even}, \citenamefont {Marks}, \citenamefont {Navrotsky}, \citenamefont
  {Mohite}, \citenamefont {Stoumpos},\ and\ \citenamefont
  {Kanatzidis}}]{bamai_n6n7_2018}%
  \BibitemOpen
  \bibfield  {author} {\bibinfo {author} {\bibfnamefont {C.~M.~M.}\
  \bibnamefont {Soe}}, \bibinfo {author} {\bibfnamefont {G.~P.}\ \bibnamefont
  {Nagabhushana}}, \bibinfo {author} {\bibfnamefont {R.}~\bibnamefont
  {Shivaramaiah}}, \bibinfo {author} {\bibfnamefont {H.}~\bibnamefont {Tsai}},
  \bibinfo {author} {\bibfnamefont {W.}~\bibnamefont {Nie}}, \bibinfo {author}
  {\bibfnamefont {J.-C.}\ \bibnamefont {Blancon}}, \bibinfo {author}
  {\bibfnamefont {F.}~\bibnamefont {Melkonyan}}, \bibinfo {author}
  {\bibfnamefont {D.~H.}\ \bibnamefont {Cao}}, \bibinfo {author} {\bibfnamefont
  {B.}~\bibnamefont {Traor{\'{e}}}}, \bibinfo {author} {\bibfnamefont
  {L.}~\bibnamefont {Pedesseau}}, \bibinfo {author} {\bibfnamefont
  {M.}~\bibnamefont {Kepenekian}}, \bibinfo {author} {\bibfnamefont
  {C.}~\bibnamefont {Katan}}, \bibinfo {author} {\bibfnamefont
  {J.}~\bibnamefont {Even}}, \bibinfo {author} {\bibfnamefont {T.~J.}\
  \bibnamefont {Marks}}, \bibinfo {author} {\bibfnamefont {A.}~\bibnamefont
  {Navrotsky}}, \bibinfo {author} {\bibfnamefont {A.~D.}\ \bibnamefont
  {Mohite}}, \bibinfo {author} {\bibfnamefont {C.~C.}\ \bibnamefont
  {Stoumpos}},\ and\ \bibinfo {author} {\bibfnamefont {M.~G.}\ \bibnamefont
  {Kanatzidis}},\ }\bibfield  {title} {\bibinfo {title} {Structural and
  thermodynamic limits of layer thickness in 2d halide perovskites},\ }\href
  {https://doi.org/10.1073/pnas.1811006115} {\bibfield  {journal} {\bibinfo
  {journal} {Proceedings of the National Academy of Sciences}\ }\textbf
  {\bibinfo {volume} {116}},\ \bibinfo {pages} {58} (\bibinfo {year}
  {2018})}\BibitemShut {NoStop}%
\bibitem [{\citenamefont {Seeck}\ \emph {et~al.}(2012)\citenamefont {Seeck},
  \citenamefont {Deiter}, \citenamefont {Pflaum}, \citenamefont {Bertram},
  \citenamefont {Beerlink}, \citenamefont {Franz}, \citenamefont {Horbach},
  \citenamefont {Schulte-Schrepping}, \citenamefont {Murphy}, \citenamefont
  {Greve},\ and\ \citenamefont {Magnussen}}]{2011petraIII}%
  \BibitemOpen
  \bibfield  {author} {\bibinfo {author} {\bibfnamefont {O.~H.}\ \bibnamefont
  {Seeck}}, \bibinfo {author} {\bibfnamefont {C.}~\bibnamefont {Deiter}},
  \bibinfo {author} {\bibfnamefont {K.}~\bibnamefont {Pflaum}}, \bibinfo
  {author} {\bibfnamefont {F.}~\bibnamefont {Bertram}}, \bibinfo {author}
  {\bibfnamefont {A.}~\bibnamefont {Beerlink}}, \bibinfo {author}
  {\bibfnamefont {H.}~\bibnamefont {Franz}}, \bibinfo {author} {\bibfnamefont
  {J.}~\bibnamefont {Horbach}}, \bibinfo {author} {\bibfnamefont
  {H.}~\bibnamefont {Schulte-Schrepping}}, \bibinfo {author} {\bibfnamefont
  {B.~M.}\ \bibnamefont {Murphy}}, \bibinfo {author} {\bibfnamefont
  {M.}~\bibnamefont {Greve}},\ and\ \bibinfo {author} {\bibfnamefont
  {O.}~\bibnamefont {Magnussen}},\ }\bibfield  {title} {\bibinfo {title} {{The
  high-resolution diffraction beamline P08 at~PETRA III}},\ }\href
  {https://doi.org/10.1107/S0909049511047236} {\bibfield  {journal} {\bibinfo
  {journal} {Journal of Synchrotron Radiation}\ }\textbf {\bibinfo {volume}
  {19}},\ \bibinfo {pages} {30} (\bibinfo {year} {2012})}\BibitemShut {NoStop}%
\bibitem [{\citenamefont {Newville}\ \emph {et~al.}(2014)\citenamefont
  {Newville}, \citenamefont {Stensitzki}, \citenamefont {Allen},\ and\
  \citenamefont {Ingargiola}}]{lmfit2014}%
  \BibitemOpen
  \bibfield  {author} {\bibinfo {author} {\bibfnamefont {M.}~\bibnamefont
  {Newville}}, \bibinfo {author} {\bibfnamefont {T.}~\bibnamefont
  {Stensitzki}}, \bibinfo {author} {\bibfnamefont {D.~B.}\ \bibnamefont
  {Allen}},\ and\ \bibinfo {author} {\bibfnamefont {A.}~\bibnamefont
  {Ingargiola}},\ }\href {https://doi.org/10.5281/ZENODO.11813} {\bibinfo
  {title} {Lmfit: Non-linear least-square minimization and curve-fitting for
  python}} (\bibinfo {year} {2014})\BibitemShut {NoStop}%
\bibitem [{\citenamefont {Huang}\ \emph {et~al.}(2021)\citenamefont {Huang},
  \citenamefont {Jamonnak}, \citenamefont {Zhao}, \citenamefont {Wang},
  \citenamefont {Hoai}, \citenamefont {Yager},\ and\ \citenamefont
  {Xu}}]{Huang2021InteractiveVS}%
  \BibitemOpen
  \bibfield  {author} {\bibinfo {author} {\bibfnamefont {X.}~\bibnamefont
  {Huang}}, \bibinfo {author} {\bibfnamefont {S.}~\bibnamefont {Jamonnak}},
  \bibinfo {author} {\bibfnamefont {Y.}~\bibnamefont {Zhao}}, \bibinfo {author}
  {\bibfnamefont {B.}~\bibnamefont {Wang}}, \bibinfo {author} {\bibfnamefont
  {M.}~\bibnamefont {Hoai}}, \bibinfo {author} {\bibfnamefont {K.~G.}\
  \bibnamefont {Yager}},\ and\ \bibinfo {author} {\bibfnamefont
  {W.}~\bibnamefont {Xu}},\ }\bibfield  {title} {\bibinfo {title} {Interactive
  visual study of multiple attributes learning model of x-ray scattering
  images},\ }\href@noop {} {\bibfield  {journal} {\bibinfo  {journal} {IEEE
  Transactions on Visualization and Computer Graphics}\ }\textbf {\bibinfo
  {volume} {27}},\ \bibinfo {pages} {1312} (\bibinfo {year}
  {2021})}\BibitemShut {NoStop}%
\bibitem [{\citenamefont {Jha}\ \emph {et~al.}(2019)\citenamefont {Jha},
  \citenamefont {Kusne}, \citenamefont {Al-Bahrani}, \citenamefont {Nguyen},
  \citenamefont {Liao}, \citenamefont {Choudhary},\ and\ \citenamefont
  {Agrawal}}]{Jha2019}%
  \BibitemOpen
  \bibfield  {author} {\bibinfo {author} {\bibfnamefont {D.}~\bibnamefont
  {Jha}}, \bibinfo {author} {\bibfnamefont {A.~G.}\ \bibnamefont {Kusne}},
  \bibinfo {author} {\bibfnamefont {R.}~\bibnamefont {Al-Bahrani}}, \bibinfo
  {author} {\bibfnamefont {N.}~\bibnamefont {Nguyen}}, \bibinfo {author}
  {\bibfnamefont {W.}~\bibnamefont {Liao}}, \bibinfo {author} {\bibfnamefont
  {A.}~\bibnamefont {Choudhary}},\ and\ \bibinfo {author} {\bibfnamefont
  {A.}~\bibnamefont {Agrawal}},\ }\bibfield  {title} {\bibinfo {title} {Peak
  area detection network for directly learning phase regions from raw x-ray
  diffraction patterns},\ }in\ \href
  {https://doi.org/10.1109/ijcnn.2019.8852096} {\emph {\bibinfo {booktitle}
  {2019 International Joint Conference on Neural Networks ({IJCNN})}}}\
  (\bibinfo  {publisher} {{IEEE}},\ \bibinfo {year} {2019})\BibitemShut
  {NoStop}%
\bibitem [{\citenamefont {Baker}\ \emph {et~al.}(2010)\citenamefont {Baker},
  \citenamefont {Jimison}, \citenamefont {Mannsfeld}, \citenamefont {Volkman},
  \citenamefont {Yin}, \citenamefont {Subramanian}, \citenamefont {Salleo},
  \citenamefont {Alivisatos},\ and\ \citenamefont {Toney}}]{Baker2010}%
  \BibitemOpen
  \bibfield  {author} {\bibinfo {author} {\bibfnamefont {J.~L.}\ \bibnamefont
  {Baker}}, \bibinfo {author} {\bibfnamefont {L.~H.}\ \bibnamefont {Jimison}},
  \bibinfo {author} {\bibfnamefont {S.}~\bibnamefont {Mannsfeld}}, \bibinfo
  {author} {\bibfnamefont {S.}~\bibnamefont {Volkman}}, \bibinfo {author}
  {\bibfnamefont {S.}~\bibnamefont {Yin}}, \bibinfo {author} {\bibfnamefont
  {V.}~\bibnamefont {Subramanian}}, \bibinfo {author} {\bibfnamefont
  {A.}~\bibnamefont {Salleo}}, \bibinfo {author} {\bibfnamefont {A.~P.}\
  \bibnamefont {Alivisatos}},\ and\ \bibinfo {author} {\bibfnamefont {M.~F.}\
  \bibnamefont {Toney}},\ }\bibfield  {title} {\bibinfo {title} {Quantification
  of thin film crystallographic orientation using x-ray diffraction with an
  area detector},\ }\href {https://doi.org/10.1021/la904840q} {\bibfield
  {journal} {\bibinfo  {journal} {Langmuir}\ }\textbf {\bibinfo {volume}
  {26}},\ \bibinfo {pages} {9146} (\bibinfo {year} {2010})}\BibitemShut
  {NoStop}%
\bibitem [{\citenamefont {Stark}(2000)}]{clahe2000}%
  \BibitemOpen
  \bibfield  {author} {\bibinfo {author} {\bibfnamefont {J.~A.}\ \bibnamefont
  {Stark}},\ }\bibfield  {title} {\bibinfo {title} {Adaptive image contrast
  enhancement using generalizations of histogram equalization},\ }\href@noop {}
  {\bibfield  {journal} {\bibinfo  {journal} {IEEE Transactions on image
  processing}\ }\textbf {\bibinfo {volume} {9}},\ \bibinfo {pages} {889}
  (\bibinfo {year} {2000})}\BibitemShut {NoStop}%
\bibitem [{\citenamefont {He}\ \emph {et~al.}(2016)\citenamefont {He},
  \citenamefont {Zhang}, \citenamefont {Ren},\ and\ \citenamefont
  {Sun}}]{resnets2016}%
  \BibitemOpen
  \bibfield  {author} {\bibinfo {author} {\bibfnamefont {K.}~\bibnamefont
  {He}}, \bibinfo {author} {\bibfnamefont {X.}~\bibnamefont {Zhang}}, \bibinfo
  {author} {\bibfnamefont {S.}~\bibnamefont {Ren}},\ and\ \bibinfo {author}
  {\bibfnamefont {J.}~\bibnamefont {Sun}},\ }\bibfield  {title} {\bibinfo
  {title} {Deep residual learning for image recognition},\ }in\ \href@noop {}
  {\emph {\bibinfo {booktitle} {Proceedings of the IEEE conference on computer
  vision and pattern recognition}}}\ (\bibinfo {year} {2016})\ pp.\ \bibinfo
  {pages} {770--778}\BibitemShut {NoStop}%
\bibitem [{\citenamefont {Perlin}(1985)}]{perlin1985}%
  \BibitemOpen
  \bibfield  {author} {\bibinfo {author} {\bibfnamefont {K.}~\bibnamefont
  {Perlin}},\ }\bibfield  {title} {\bibinfo {title} {An image synthesizer},\
  }\href {https://doi.org/10.1145/325165.325247} {\bibfield  {journal}
  {\bibinfo  {journal} {SIGGRAPH Comput. Graph.}\ }\textbf {\bibinfo {volume}
  {19}},\ \bibinfo {pages} {287–296} (\bibinfo {year} {1985})}\BibitemShut
  {NoStop}%
\bibitem [{\citenamefont {Paszke}\ \emph {et~al.}(2019)\citenamefont {Paszke},
  \citenamefont {Gross}, \citenamefont {Massa}, \citenamefont {Lerer},
  \citenamefont {Bradbury}, \citenamefont {Chanan}, \citenamefont {Killeen},
  \citenamefont {Lin}, \citenamefont {Gimelshein}, \citenamefont {Antiga},
  \citenamefont {Desmaison}, \citenamefont {Kopf}, \citenamefont {Yang},
  \citenamefont {DeVito}, \citenamefont {Raison}, \citenamefont {Tejani},
  \citenamefont {Chilamkurthy}, \citenamefont {Steiner}, \citenamefont {Fang},
  \citenamefont {Bai},\ and\ \citenamefont {Chintala}}]{Pytorch2019}%
  \BibitemOpen
  \bibfield  {author} {\bibinfo {author} {\bibfnamefont {A.}~\bibnamefont
  {Paszke}}, \bibinfo {author} {\bibfnamefont {S.}~\bibnamefont {Gross}},
  \bibinfo {author} {\bibfnamefont {F.}~\bibnamefont {Massa}}, \bibinfo
  {author} {\bibfnamefont {A.}~\bibnamefont {Lerer}}, \bibinfo {author}
  {\bibfnamefont {J.}~\bibnamefont {Bradbury}}, \bibinfo {author}
  {\bibfnamefont {G.}~\bibnamefont {Chanan}}, \bibinfo {author} {\bibfnamefont
  {T.}~\bibnamefont {Killeen}}, \bibinfo {author} {\bibfnamefont
  {Z.}~\bibnamefont {Lin}}, \bibinfo {author} {\bibfnamefont {N.}~\bibnamefont
  {Gimelshein}}, \bibinfo {author} {\bibfnamefont {L.}~\bibnamefont {Antiga}},
  \bibinfo {author} {\bibfnamefont {A.}~\bibnamefont {Desmaison}}, \bibinfo
  {author} {\bibfnamefont {A.}~\bibnamefont {Kopf}}, \bibinfo {author}
  {\bibfnamefont {E.}~\bibnamefont {Yang}}, \bibinfo {author} {\bibfnamefont
  {Z.}~\bibnamefont {DeVito}}, \bibinfo {author} {\bibfnamefont
  {M.}~\bibnamefont {Raison}}, \bibinfo {author} {\bibfnamefont
  {A.}~\bibnamefont {Tejani}}, \bibinfo {author} {\bibfnamefont
  {S.}~\bibnamefont {Chilamkurthy}}, \bibinfo {author} {\bibfnamefont
  {B.}~\bibnamefont {Steiner}}, \bibinfo {author} {\bibfnamefont
  {L.}~\bibnamefont {Fang}}, \bibinfo {author} {\bibfnamefont {J.}~\bibnamefont
  {Bai}},\ and\ \bibinfo {author} {\bibfnamefont {S.}~\bibnamefont
  {Chintala}},\ }\bibfield  {title} {\bibinfo {title} {Pytorch: An imperative
  style, high-performance deep learning library},\ }in\ \href
  {https://proceedings.neurips.cc/paper/2019/file/bdbca288fee7f92f2bfa9f7012727740-Paper.pdf}
  {\emph {\bibinfo {booktitle} {Advances in Neural Information Processing
  Systems}}},\ Vol.~\bibinfo {volume} {32},\ \bibinfo {editor} {edited by\
  \bibinfo {editor} {\bibfnamefont {H.}~\bibnamefont {Wallach}}, \bibinfo
  {editor} {\bibfnamefont {H.}~\bibnamefont {Larochelle}}, \bibinfo {editor}
  {\bibfnamefont {A.}~\bibnamefont {Beygelzimer}}, \bibinfo {editor}
  {\bibfnamefont {F.}~\bibnamefont {d\textquotesingle Alch\'{e}-Buc}}, \bibinfo
  {editor} {\bibfnamefont {E.}~\bibnamefont {Fox}},\ and\ \bibinfo {editor}
  {\bibfnamefont {R.}~\bibnamefont {Garnett}}}\ (\bibinfo  {publisher} {Curran
  Associates, Inc.},\ \bibinfo {year} {2019})\BibitemShut {NoStop}%
\bibitem [{\citenamefont {Loshchilov}\ and\ \citenamefont
  {Hutter}(2019)}]{loshchilov2019decoupled}%
  \BibitemOpen
  \bibfield  {author} {\bibinfo {author} {\bibfnamefont {I.}~\bibnamefont
  {Loshchilov}}\ and\ \bibinfo {author} {\bibfnamefont {F.}~\bibnamefont
  {Hutter}},\ }\bibfield  {title} {\bibinfo {title} {Decoupled weight decay
  regularization},\ }in\ \href {https://openreview.net/forum?id=Bkg6RiCqY7}
  {\emph {\bibinfo {booktitle} {International Conference on Learning
  Representations}}}\ (\bibinfo {year} {2019})\BibitemShut {NoStop}%
\bibitem [{\citenamefont {Ioffe}\ and\ \citenamefont
  {Szegedy}(2015)}]{batch_normalization_2015}%
  \BibitemOpen
  \bibfield  {author} {\bibinfo {author} {\bibfnamefont {S.}~\bibnamefont
  {Ioffe}}\ and\ \bibinfo {author} {\bibfnamefont {C.}~\bibnamefont
  {Szegedy}},\ }\bibfield  {title} {\bibinfo {title} {Batch normalization:
  Accelerating deep network training by reducing internal covariate shift},\
  }in\ \href {https://proceedings.mlr.press/v37/ioffe15.html} {\emph {\bibinfo
  {booktitle} {Proceedings of the 32nd International Conference on Machine
  Learning}}},\ \bibinfo {series} {Proceedings of Machine Learning Research},
  Vol.~\bibinfo {volume} {37},\ \bibinfo {editor} {edited by\ \bibinfo {editor}
  {\bibfnamefont {F.}~\bibnamefont {Bach}}\ and\ \bibinfo {editor}
  {\bibfnamefont {D.}~\bibnamefont {Blei}}}\ (\bibinfo  {publisher} {PMLR},\
  \bibinfo {address} {Lille, France},\ \bibinfo {year} {2015})\ pp.\ \bibinfo
  {pages} {448--456}\BibitemShut {NoStop}%
\bibitem [{\citenamefont {Neubeck}\ and\ \citenamefont
  {Van~Gool}(2006)}]{nms2006}%
  \BibitemOpen
  \bibfield  {author} {\bibinfo {author} {\bibfnamefont {A.}~\bibnamefont
  {Neubeck}}\ and\ \bibinfo {author} {\bibfnamefont {L.}~\bibnamefont
  {Van~Gool}},\ }\bibfield  {title} {\bibinfo {title} {Efficient non-maximum
  suppression},\ }in\ \href@noop {} {\emph {\bibinfo {booktitle} {18th
  International Conference on Pattern Recognition (ICPR'06)}}},\ Vol.~\bibinfo
  {volume} {3}\ (\bibinfo {organization} {IEEE},\ \bibinfo {year} {2006})\ pp.\
  \bibinfo {pages} {850--855}\BibitemShut {NoStop}%
\bibitem [{\citenamefont {Wu}\ \emph {et~al.}(2005)\citenamefont {Wu},
  \citenamefont {Otoo},\ and\ \citenamefont
  {Shoshani}}]{label_segmentation_2005}%
  \BibitemOpen
  \bibfield  {author} {\bibinfo {author} {\bibfnamefont {K.}~\bibnamefont
  {Wu}}, \bibinfo {author} {\bibfnamefont {E.}~\bibnamefont {Otoo}},\ and\
  \bibinfo {author} {\bibfnamefont {A.}~\bibnamefont {Shoshani}},\ }\bibfield
  {title} {\bibinfo {title} {Optimizing connected component labeling
  algorithms},\ }in\ \href {https://doi.org/10.1117/12.596105} {\emph {\bibinfo
  {booktitle} {{SPIE} Proceedings}}},\ \bibinfo {editor} {edited by\ \bibinfo
  {editor} {\bibfnamefont {J.~M.}\ \bibnamefont {Fitzpatrick}}\ and\ \bibinfo
  {editor} {\bibfnamefont {J.~M.}\ \bibnamefont {Reinhardt}}}\ (\bibinfo
  {publisher} {{SPIE}},\ \bibinfo {year} {2005})\BibitemShut {NoStop}%
\bibitem [{\citenamefont {van~der Walt}\ \emph {et~al.}(2014)\citenamefont
  {van~der Walt}, \citenamefont {Sch\"{o}nberger}, \citenamefont
  {Nunez-Iglesias}, \citenamefont {Boulogne}, \citenamefont {Warner},
  \citenamefont {Yager}, \citenamefont {Gouillart},\ and\ \citenamefont
  {Yu}}]{scikit_image_2014}%
  \BibitemOpen
  \bibfield  {author} {\bibinfo {author} {\bibfnamefont {S.}~\bibnamefont
  {van~der Walt}}, \bibinfo {author} {\bibfnamefont {J.~L.}\ \bibnamefont
  {Sch\"{o}nberger}}, \bibinfo {author} {\bibfnamefont {J.}~\bibnamefont
  {Nunez-Iglesias}}, \bibinfo {author} {\bibfnamefont {F.}~\bibnamefont
  {Boulogne}}, \bibinfo {author} {\bibfnamefont {J.~D.}\ \bibnamefont
  {Warner}}, \bibinfo {author} {\bibfnamefont {N.}~\bibnamefont {Yager}},
  \bibinfo {author} {\bibfnamefont {E.}~\bibnamefont {Gouillart}},\ and\
  \bibinfo {author} {\bibfnamefont {T.}~\bibnamefont {Yu}},\ }\bibfield
  {title} {\bibinfo {title} {scikit-image: image processing in python},\ }\href
  {https://doi.org/10.7717/peerj.453} {\bibfield  {journal} {\bibinfo
  {journal} {{PeerJ}}\ }\textbf {\bibinfo {volume} {2}},\ \bibinfo {pages}
  {e453} (\bibinfo {year} {2014})}\BibitemShut {NoStop}%
\bibitem [{\citenamefont {Byrd}\ \emph {et~al.}(1995)\citenamefont {Byrd},
  \citenamefont {Lu}, \citenamefont {Nocedal},\ and\ \citenamefont
  {Zhu}}]{L_BFGS_B_1995}%
  \BibitemOpen
  \bibfield  {author} {\bibinfo {author} {\bibfnamefont {R.~H.}\ \bibnamefont
  {Byrd}}, \bibinfo {author} {\bibfnamefont {P.}~\bibnamefont {Lu}}, \bibinfo
  {author} {\bibfnamefont {J.}~\bibnamefont {Nocedal}},\ and\ \bibinfo {author}
  {\bibfnamefont {C.}~\bibnamefont {Zhu}},\ }\bibfield  {title} {\bibinfo
  {title} {A limited memory algorithm for bound constrained optimization},\
  }\href {https://doi.org/10.1137/0916069} {\bibfield  {journal} {\bibinfo
  {journal} {{SIAM} Journal on Scientific Computing}\ }\textbf {\bibinfo
  {volume} {16}},\ \bibinfo {pages} {1190} (\bibinfo {year}
  {1995})}\BibitemShut {NoStop}%
\end{thebibliography}%

\end{document}